\documentclass[letterpaper]{article} 
\usepackage{aaai23}  
\usepackage{times}  
\usepackage{helvet}  
\usepackage{courier}  
\usepackage[hyphens]{url}  
\usepackage{graphicx} 
\usepackage{natbib}  
\usepackage{caption} 
\usepackage{algorithm}
\usepackage{algpseudocode}
\usepackage{multirow}
\usepackage{wrapfig}
\usepackage{amsmath,amssymb} 
\usepackage{color}
\usepackage{newfloat}
\usepackage{listings}

\usepackage[capitalize]{cleveref}
\crefname{section}{Sec.}{Secs.}
\Crefname{section}{Section}{Sections}
\Crefname{table}{Table}{Tables}
\crefname{table}{Tab.}{Tabs.}
\crefname{algocf}{alg.}{algs.}

\DeclareCaptionStyle{ruled}{labelfont=normalfont,labelsep=colon,strut=off} 
\floatstyle{ruled}
\newfloat{listing}{tb}{lst}{}
\floatname{listing}{Listing}
\title{Defending Black-box Skeleton-based Human Activity Classifiers}
\author{
He Wang\equalcontrib\footnote{corresponding author, h.e.wang@leeds.ac.uk}\textsuperscript{\rm 1},
Yunfeng Diao\equalcontrib\textsuperscript{\rm 2},
Zichang Tan\textsuperscript{\rm 3},
Guodong Guo\textsuperscript{\rm 3}
}
\affiliations {
    \textsuperscript{\rm 1} University of Leeds,  UK 
    \textsuperscript{\rm 2} Hefei University of Technology, Hefei China\\
    \textsuperscript{\rm 3 }Institute of Deep Learning, Baidu Research, Beijing China
    
}

\usepackage{bibentry}

\begin{document}

\maketitle

\begin{abstract}
Skeletal motions have been heavily relied upon for human activity recognition (HAR). Recently, a universal vulnerability of skeleton-based HAR has been identified across a variety of classifiers and data, calling for mitigation. To this end, we propose the first black-box defense method for skeleton-based HAR to our best knowledge. Our method is featured by full Bayesian treatments of the clean data, the adversaries and the classifier, leading to (1) a new Bayesian Energy-based formulation of robust discriminative classifiers, (2) a new adversary sampling scheme based on natural motion manifolds, and (3) a new post-train Bayesian strategy for black-box defense. We name our framework Bayesian Energy-based Adversarial Training or BEAT. BEAT is straightforward but elegant, which turns vulnerable black-box classifiers into robust ones without sacrificing accuracy. It demonstrates surprising and universal effectiveness across a wide range of skeletal HAR classifiers and datasets, under various attacks. Code is available at https://github.com/realcrane/Defending-Black-box-Skeleton-based-Human-Activity-Classifiers.
\end{abstract}

\section{Introduction}
Classification is a fundamental task where deep learning has achieved the state-of-the-art performance. However, deep learning models are vulnerable to strategically computed perturbations on the inputs, a.k.a. adversarial attack~\cite{chakraborty_adversarial_2018}. The universality of the vulnerability has caused alarming concerns because the perturbations are imperceptible to humans but destructive to machine intelligence. Subsequently, defense methods have emerged as a new field recently~\cite{chakraborty_adversarial_2018} most of which are focused on static data~\cite{akhtar_advances_2021}. More recently, adversarial attack has started to appear on time-series data~\cite{karim_adversarial_2020,wei2019sparse,liu_adversarial_2020}, but the corresponding defense research has been largely unexplored, especially for skeleton-based HAR, an important time-series data where a universal vulnerability has been recently identified~\cite{wang_understanding_2021,diao_basarblack-box_2021}, urgently calling for mitigation. To this end, we fill this gap by proposing a new defense framework based on adversarial training (AT). 

Defense on HAR presents several challenges. First, while most AT methods seek to resist attacks from the most aggressive adversarial sample~\cite{silva_opportunities_2020}, the whole adversarial distribution should be considered~\cite{ye_bayesian_2018,dong2020adversarial}. But this has been mainly done on images assuming a simple structure of the adversarial distribution (e.g. Gaussian). How to model the adversarial distribution of skeletal motions has not been explored. Further, adversarial samples are near or on the data manifold constrained by the motion dynamics~\cite{wang_understanding_2021,diao_basarblack-box_2021} which is different different from other time-series data, e.g. videos. Naive adaptation of existing AT methods~\cite{madry_towards_2018,MART,TRADES}, e.g. image-based, leads to crude approximation of this manifold, resulting in either ineffective defense or compromise of classification accuracy. While mitigation is possible for images~\cite{miyato2018virtual,carmon2019unlabeled,Songlocal}, it is under-explored for skeletal motions. Finally, most AT methods try to find the best model that can resist attacks. From the Bayesian perspective, this is a \textit{point estimation} on the model. An ensemble of different models can provide better robustness~\cite{ye_bayesian_2018,Bortolussi_onTheRobustness_2022} but how to find them for skeletal motions has not been explored.  

To address the challenges, we present a new framework for robust skeleton-based HAR. Extending Energy-based methods~\cite{StochasticSecurity,zhu2021towards,Lee2020AdversarialTO} which model the data distribution $p(\mathbf{x}, \mathbf{y})$ of data $\mathbf{x}$ and labels $\mathbf{y}$, and sample the adversaries $\Tilde{\mathbf{x}}$ based on simplified adversarial distributions, we consider the full adversarial distribution by jointly modeling $\mathbf{x}$, $\mathbf{y}$, $\Tilde{\mathbf{x}}$ and the classifier parameterized by $\theta$: $p(\mathbf{x}, \mathbf{y}, \Tilde{\mathbf{x}},\theta)$, to give a full Bayesian treatment on all relevant factors. This involves new Bayesian treatments on both $\Tilde{\mathbf{x}}$ and $\theta$. 

We first re-interpret discriminative classifiers into an energy-based model that maximizes the joint probability $p(\mathbf{x}, \mathbf{y})$~\cite{grathwohl_your_2020}. Then one key novelty is we assume we can observe all adversarial samples $\Tilde{\mathbf{x}}$ during training, so that we can model the clean-adversarial joint probability $p(\mathbf{x}, \Tilde{\mathbf{x}}, \mathbf{y})$. This joint probability considers the full adversarial distribution $p(\Tilde{\mathbf{x}} | \mathbf{x}, \mathbf{y})$ conditioned on the clean data where we propose a new adversary sampling scheme along the natural motion manifold. Overall, this leads to a more general adversarial distribution parameterization than existing methods which rely on a pre-defined attacker~\cite{madry_towards_2018} or isotropic distributions~\cite{lecuyer_certified_2019} for adversary sampling. Further, we incorporate classifier parameters $\theta$ into the joint probability to explore the full space of robust classifiers. However, different from existing Bayesian Neural Networks (BNNs) research which seeks to turn the classifier into a BNN, we keep the classifier intact and append small extra Bayesian components to achieve robustness. This makes our method applicable to pre-trained classifiers and requires minimal prior knowledge of them, achieving black-box defense.

As a result, we propose a new joint Bayesian perspective on the clean data, the adversarial samples and the classifier. We name our method Bayesian Energy-based Adversarial Training (BEAT). BEAT can turn pre-trained classifiers into resilient ones. It also circumvents model re-training, avoids heavy memory footprint and speeds up adversarial training. BEAT leads to a more general defense mechanism against a variety of attackers which are not known \textit{a priori} during training. We evaluate BEAT on several state-of-the-art classifiers across a number of benchmark datasets, and compare it with existing methods. Overall, BEAT can effectively boost the robustness of classifiers against attack. Empirical results show that BEAT does not severely sacrifice accuracy for robustness, as opposed to the common observation of such trade-off in other AT methods~\cite{Yangcloserlook}.

Formally, we propose: 1) the first black-box defense method for skeleton-based HAR to our best knowledge. 2) a new Bayesian perspective on a joint distribution of normal data, adversarial samples and classifiers. 3) a new post-train Bayesian strategy to keep the blackboxness of classifiers and avoid heavy memory footprint. 

\section{Related Work}
\subsection{Adversarial Attack}
Since the vulnerability of deep learning was identified~\cite{GoodfellowFGSM}, the community has developed diverse adversarial attacks on different data types, e.g. texts~\cite{liang2018text}, graphs~\cite{dai2018adversarial,zugner2018adversarial} and physical objects~\cite{evtimov2017robust,athalye2018synthesizing}. While static data has attracted most of the attention, the attack on time-series data has recently emerged~\cite{karim_adversarial_2020,wei2019sparse}. One active sub-field is Human Activity Recognition. Unlike other data, motion has unique features such as dynamics and human body topology~\cite{wei2020heuristic,diao_basarblack-box_2021}, which makes it difficult to adapt generic methods~\cite{BA,GoodfellowFGSM,CW2017}. Therefore, existing HAR attacks are designed for specific data types. Adversaries have been developed in video-based recognition~\cite{zhang2020motion,pony2021over,hwang2021just,wei2020heuristic} and multi-modal setting~\cite{kumar2020finding}. Very recently, skeleton-based HAR has been shown to be extremely vulnerable~\cite{liu_adversarial_2020,tanaka2021adversarial,wang_understanding_2021,diao_basarblack-box_2021,zheng_towards_2020}. Adversarial examples can be generated by Generative Adversarial Networks~\cite{liu_adversarial_2020}, optimization based on a new perceptual metric~\cite{wang_understanding_2021}, or exploring the interplay between the classification boundary and the natural motion manifold under the hard-label black-box setting~\cite{diao_basarblack-box_2021}. Orthogonal to attack, BEAT propose a new defense framework for skeleton-based HAR to address the urgent challenges.

\subsection{Adversarial Training}
AT methods~\cite{bai2021recent,GoodfellowFGSM,madry_towards_2018,MART,TRADES} are among the most effective defense techniques to date. They train the classifier to resist the attack from the most aggressive adversarial examples. However, these specific adversarial examples may not sufficiently represent the whole adversarial sample distribution, leading to difficulties when facing unseen and stronger adversaries~\cite{uesato2018adversarial,song2018improving}. Further, existing AT methods all compromise the standard accuracy to a certain extent~\cite{Yangcloserlook}. The study of generalization in AT is still under-explored.

Stutz et al.~\cite{stutz2020confidence} proposed to reject the unseen attacks by reducing the confidence scores of adversarial examples, while Poursaeed et al.~\cite{poursaeed2021robustness} generated diversified adversarial changes in the examples used in AT. Learning the adversarial sample distribution is shown to improve the robustness~\cite{ye_bayesian_2018}. Dong et al.~\cite{dong2020adversarial} extended adversarial training through explicitly or implicitly modeling the adversarial distributions. However, they are designed for static image data. A key difference between their methods and ours is that we consider a joint distribution of the normal data, adversarial samples and the classifier, which allows us to design an adversary sampling scheme for skeleton motions and fully explore the space of robust classifiers.

For generalization, early studies~\cite{tsipras2018robustness,zhang2019theoretically} postulate that there should be an inherent tradeoff between standard accuracy and adversarial robustness. However, recent works challenge this postulation. Stutz et al.~\cite{stutz2019disentangling} reckoned that the adversarial robustness in the underlying natural manifold is related to generalization. Empirically, robust semi/unsupervised training~\cite{miyato2018virtual,carmon2019unlabeled} utilizing extra data can mitigate this problem. The trade-off could in theory be eliminated under the infinite data assumption~\cite{raghunathan2020understanding}. Yang et al.~\cite{Yangcloserlook} showed some image datasets are distributionally separate, indicating there exists an ideal robust classifier that does not compromise the accuracy. To our best knowledge, we propose the first skeleton-based HAR black-box defense which demonstrates the existence of a resilient classifier without the inherent accuracy-robustness trade-off. 


\section{Methodology}
\subsection{Background in Energy-based Models}
Given data $\mathbf{x} \in \mathbf{X}$ and label $\mathbf{y}$, a discriminative classifier can be generalized from an energy perspective by modeling the joint distribution $p_{\theta}(\mathbf{x}, y) = \frac{exp(g_{\theta}(\mathbf{x})[y])}{Z(\theta)}$ where $y\in\mathbf{y}$ and $\theta$ is the model parameters~\cite{grathwohl_your_2020}. Since $p_{\theta}(\mathbf{x}, y) = p_{\theta}(y|\mathbf{x})p_{\theta}(\mathbf{x})$ and $p_{\theta}(y|\mathbf{x})$ is what classifiers maximize, the key difference is $p_{\theta}(\mathbf{x})$ which can be parameterized by an energy function:
\begin{equation}
\label{eq:dataDis}
    p_{\theta}(\mathbf{x}) = \frac{exp(-E_{\theta}(\mathbf{x}))}{Z(\theta)} = \frac{\sum_{y\in\mathbf{y}} exp(g_{\theta}(\mathbf{x})[y])}{Z(\theta)}
\end{equation}
where $E_{\theta}$ is an energy function parameterized by $\theta$, $Z(\theta) = \int_{\mathbf x} exp(-E_\theta({\mathbf x}))d{\mathbf x}$ is a normalizing constant. This energy-based interpretation allows an arbitrary $E_{\theta}$ to describe a continuous density function, as long as it assigns low energy values to observations and high energy everywhere else. This leads to a generalization of discriminative classifiers: $E$ can be an exponential function as shown in \cref{eq:dataDis} where $g_{\theta}$ is a classifier and $g_{\theta}(\mathbf{x})[y]$ gives the $y$th logit for class $y$. $\theta$ can be learned via maximizing the log likelihood:
\begin{align}
\label{eq:jointEnergy}
    &log\ p_{\theta}(\mathbf{x}, y) = log\ p_{\theta}(y|\mathbf{x}) + log\ p_{\theta}(\mathbf{x}) \nonumber \text{  where} \\
    &p_{\theta}(y|\mathbf{x}) = \frac{p_{\theta}(\mathbf{x},y)}{p_{\theta}(\mathbf{x})} = \frac{exp(g_{\theta}(\mathbf{x})[y])}{\sum_{y'\in\mathbf{y}} exp(g_{\theta}(\mathbf{x})[y'])}
\end{align}
Compared with only maximizing $log\ p(y|\mathbf{x})$ as discriminative classifiers do, maximizing $log\  p(\mathbf{x}, y)$ can provide many benefits such as good accuracy, robustness and out-of-distribution detection~\cite{grathwohl_your_2020}.

\subsection{Joint Distribution of Data and Adversaries}
\label{Sec.data}
A robust classifier that can resist adversarial attacks, i.e. correctly classifying both the clean $\mathbf{x}$ and the adversarial samples $\Tilde{\mathbf{x}}$, needs to consider the clean data, the adversarial samples and the attacker simultaneously:
\begin{equation}
    g_\theta(\mathbf{x}) = g_\theta(\Tilde{\mathbf{x}}) \text{ where}\ \Tilde{\mathbf{x}} = \mathbf{x} + \mathbf{\sigma}, \mathbf{\sigma}\in\pi
\end{equation}
where a classifier $g_\theta$ takes an input and outputs a class label, and $\mathbf{\sigma}$ is drawn from some perturbation set $\pi$, computed by an attacker. Since the attacker is not known \textit{a priori}, $g_\theta$ needs to capture the whole adversarial distribution to be able to resist potential attacks post-train. However, modeling the adversarial distribution is non-trivial as they are not observed during training. This has led to two strategies: defending against the most adversarial sample from an attacker (a.k.a Adversarial Training or AT~\cite{madry_towards_2018}) or train on data with noises (a.k.a Randomized Smoothing or RS~\cite{lecuyer_certified_2019}). However, both approaches lead to a trade-off between accuracy and robustness~\cite{Yangcloserlook}. We speculate that it is because neither can fully capture the adversarial distribution.

We start from a straightforward yet key conceptual deviation from literature~\cite{chakraborty_adversarial_2018}: assuming there is an adversarial distribution over all possible attackers and it can be observed during training. Although it is hard to depict the adversarial distribution directly, all adversarial samples are close to the clean data~\cite{diao_basarblack-box_2021} hence should also have relatively low energy. Therefore, we add the adversarial sample $\Tilde{\mathbf{x}}$ to the joint distribution $p(\mathbf{x}, \Tilde{\mathbf{x}}, y)$, and further extend it into a new energy-based model:
\begin{equation}
\label{eq:fullJointEnergy}
    p_{\theta}(\mathbf{x}, \Tilde{\mathbf{x}}, y) = \frac{exp\{g_{\theta}(\mathbf{x})[y] + g_{\theta}(\mathbf{\Tilde{\mathbf{x}}})[y] - \lambda d(\mathbf{x}, \Tilde{\mathbf{x}})\}}{Z(\theta)}
\end{equation}
where $\mathbf{x}$ and $\Tilde{\mathbf{x}}$ are the clean samples and their corresponding adversaries under class $y$. $\lambda$ is a weight and $d(\mathbf{x}, \Tilde{\mathbf{x}})$ measures the distance between the clean samples and their adversaries. \cref{eq:fullJointEnergy} bears two assumptions. First, adversaries should also be in the low-energy (high-density) area as they are assumed to be observed. Also, their energy should increase (or density should decrease) when they deviate away from the clean samples, governed by $d(\mathbf{x}, \Tilde{\mathbf{x}})$. 

Looking closely, $p_{\theta}(\mathbf{x}, \Tilde{\mathbf{x}}, y) = p_{\theta}(\Tilde{\mathbf{x}} | \mathbf{x}, y)p_{\theta}(\mathbf{x}, y)$, where $p_{\theta}(\mathbf{x}, y)$ is the same as in \cref{eq:jointEnergy}. $p_{\theta}(\Tilde{\mathbf{x}} | \mathbf{x}, y)$ is a new term. To further understand this term, for each data sample $\mathbf{x}$, we take a Bayesian perspective and assume there is a distribution of adversarial samples $\Tilde{\mathbf{x}}$ around $\mathbf{x}$. This assumption is reasonable as every adversarial sample can be traced back to a clean sample, and there is a \textit{one-to-many} mapping from the clean samples to the adversarial samples. Then $p_{\theta}(\Tilde{\mathbf{x}} | \mathbf{x}, y)$ is a \textit{full Bayesian treatment} of all adversarial samples:
\begin{equation}
\label{eq:adDist}
    p_{\theta}(\Tilde{\mathbf{x}} | \mathbf{x}, y) = \frac{p_{\theta}(\mathbf{x}, \Tilde{\mathbf{x}}, y)}{p_{\theta}(\mathbf{x}, y)} = exp\{g_{\theta}(\mathbf{\Tilde{\mathbf{x}}})[y] - \lambda d(\mathbf{x}, \Tilde{\mathbf{x}})\}
\end{equation}
where the intractable $Z(\theta)$ is conveniently cancelled. \cref{eq:adDist} is a key component in BEAT as it provides an energy-based parameterization, so that we are sure adversarial samples will be given low energy values and thus high density (albeit unnormalized). Through \cref{eq:adDist}, our classifier is now capable of taking the adversarial sample distribution into consideration during training.

\subsubsection{Natural Motion Manifold for AT in HAR}
$d(\mathbf{x}, \Tilde{\mathbf{x}})$ can be realized implicitly, e.g. using another model to learn the data manifold to compute the distance, but this would break BEAT into a two-stage process. Therefore, we employ explicit parameterization to achieve end-to-end training. The motion manifold is well described by the motion dynamics and bone lengths~\cite{wang_energy-driven_2015,wang_spatio-temporal_2021,wang_harmonic_2013,tang_realtime_2022}. Therefore, we design $d$ so that the energy function in \cref{eq:adDist} also assigns low energy values to the adversarial samples bearing similar motion dynamics and bone lengths:
\begin{align}
\label{eq:simMeasure}
    d(\mathbf{x}, \Tilde{\mathbf{x}}) = \frac{1}{MB}\sum||BL(\mathbf{x}) - BL(\Tilde{\mathbf{x}})||^2_p \nonumber \\
    + \frac{1}{MJ}\sum||q_{m,j}^k(\mathbf{x}) - \Tilde{q}^k_{m,j}(\Tilde{\mathbf{x}})||^2_p
\end{align}
where $\mathbf{x}$, $\Tilde{\mathbf{x}} \in \mathbb{R}^{M\times 3J}$ are motions containing a sequence of $M$ poses (frames), each of which contains $J$ 3D joint locations and $B$ bones. $BL$ computes the bone lengths in each frame. $q_{m,j}^k$ and $\Tilde{q}^k_{m,j}$ are the $k$th-order derivative of the $j$th joint in the $m$th frame in the clean sample and its adversary respectively. $k\in[0,2]$. This is because a dynamical system can be represented by a collection of derivatives at different orders. For human motions, we empirically consider the first three orders: position, velocity and acceleration. High-order information can also be considered but would incur extra computation. $||\cdot||_p$ is  the $\ell_p$ norm. We set $p=2$ but other values are also possible. Overall, the first term is a bone-length energy and the second one is motion dynamics energy. Both energy terms together defines a distance function centered at a clean data $\mathbf{x}$. This distance function helps to quantify how likely an adversarial sample near $\mathbf{x}$ is, so \cref{eq:simMeasure} describes the adversarial distribution near the motion manifold.

\subsection{Bayesian Classifier for Further Robustness}
\subsubsection{Maximum-likelihood}
One natural choice for adversarial training is to maximize~\cref{eq:fullJointEnergy}. We can learn $\theta$ by maximizing the log-likelihood of the joint probability, where $g_{\theta}$ is an arbitrary classifier:
\begin{align}
\label{eq:logLikelihood}
    log\ &p_{\theta}(\mathbf{x}, \Tilde{\mathbf{x}}, y) = log\  p_{\theta}(\Tilde{\mathbf{x}} | \mathbf{x}, y) + log\ p_{\theta}(\mathbf{x}, y) \nonumber \\
    &= log\  p_{\theta}(\Tilde{\mathbf{x}} | \mathbf{x}, y) + log\ p_{\theta}(y|\mathbf{x}) + log\ p_{\theta}(\mathbf{x}) 
\end{align}
$log\ p_{\theta}(y|\mathbf{x})$ is simply a classification likelihood and can be estimated via e.g. cross-entropy. Both $p_{\theta}(\mathbf{x})$ and $p_{\theta}(\Tilde{\mathbf{x}} | \mathbf{x}, y)$ are intractable, so sampling is needed. Then $\theta$ can be optimized using stochastic gradient methods.

\subsubsection{A Bayesian Perspective on the Classifier}
Although \cref{eq:logLikelihood} considers the full distribution of the adversarial samples, it is still a \textit{point estimation} with respect to $\theta$. From a Bayesian perspective, there is a distribution of models which can correctly classify $\mathbf{x}$, i.e. there is an infinite number of ways to draw the classification boundaries. Our inspiration is that a single boundary can be robust against certain adversaries, e.g. the distance between the boundary and some clean data samples are large hence requiring larger perturbations for attack, then a collection of boundaries can be more robust because they provide different between-class distances~\cite{Yangcloserlook} and local boundary continuity~\cite{liu_loss_2020}. Therefore, we augment~\cref{eq:fullJointEnergy} to incorporate the network weights $\theta$: 

\begin{equation}
\label{eq:BEAT}
    p(\theta, \mathbf{x}, \Tilde{\mathbf{x}}, y) = p(\mathbf{x}, \Tilde{\mathbf{x}}, y | \theta) p(\theta)
\end{equation}
where $p(\mathbf{x}, \Tilde{\mathbf{x}}, y | \theta)$ is essentially~\cref{eq:fullJointEnergy} and $p(\theta)$ is the prior of network weights. This way, we have a new Bayesian joint model of clean data, adversarial samples and the classifier. From the Bayesian perspective, \cref{eq:logLikelihood} is equivalent to using a flat $p(\theta)$ and applying iterative \textit{Maximum a posteriori} (MAP) optimization. However, even with a flat prior, a MAP optimization is still a point estimation on the model, and cannot fully utilize the full posterior distribution~\cite{saatci_bayesian_2017}. In contrast, we propose a method based on \textit{Bayesian Model Averaging}:
\begin{align}
\label{eq:bayesianClassifier}
    p(y' | \mathbf{x}', \mathbf{x}, \Tilde{\mathbf{x}}, y) = E_{\theta\sim p(\theta)}[p(y' | \mathbf{x}', \mathbf{x}, \Tilde{\mathbf{x}}, y, \theta)] \nonumber \\
    \approx \frac{1}{N}\sum_{i = 1}^Np(y'|\mathbf{x}',\theta_i), \theta \sim p(\theta | \mathbf{x}, \Tilde{\mathbf{x}}, y)
\end{align}
where $\mathbf{x}'$ and $y'$ are a new sample and its predicted label, $p(\theta)$ is a flat prior, $N$ is the number of models. We expect such a Bayesian classifier to be more robust against attack while achieving good accuracy, because models from the high probability regions of $p( \theta | \mathbf{x}, \Tilde{\mathbf{x}}, y)$ provide both. This is vital as we do not know the attacker in advance. To train such a classifier, the posterior distribution $p( \theta | \mathbf{x}, \Tilde{\mathbf{x}}, y)$ needs to be sampled as it is intractable. 



\subsubsection{Necessity of a Post-train Bayesian Strategy}

Unfortunately, it is not straightforward to design such a Bayesian treatment (\cref{eq:bayesianClassifier}) on existing classifiers due to several factors. First, sampling the posterior distribution $p( \theta | \mathbf{x}, \Tilde{\mathbf{x}}, y)$ is prohibitively slow. Considering the large number of parameters in action classifiers (commonly at least several millions), sampling would mix extremely slowly in such a high dimensional space (if mix at all). In addition, from the perspective of end-users, large models are normally pre-trained on large datasets then shared. The end-users can fine-tune or directly use the pre-trained model. It is not desirable to re-train the models. Finally, most classifiers consists of two parts: feature extraction and boundary computation. The data is pulled through the first part to be mapped into a latent feature space, then the boundary is computed, e.g. through fully-connected layers.  The feature extraction component is well learned in the pre-trained model. Keeping the features intact can avoid re-training the model, and avoid undermining other tasks when the features are learned for multiple tasks, e.g. under representation/self-supervised learning.

Therefore, we propose a \textit{post-train} BEAT for black-box defense. We keep the pre-trained classifier intact and append a tiny model with parameters $\theta'$ behind the classifier using a skip connection: logits = $f_{\theta'}(\phi(\mathbf{x})) + g_\theta(\mathbf{x})$, in contrast to the original logits = $g_\theta(\mathbf{x})$. $\phi(\mathbf{x})$ can be the latent features of $\mathbf{x}$ or the original logits $\phi(\mathbf{x}) = g_\theta(\mathbf{x})$. We employ the latter setting based on preliminary experiments (see Appendix A), and to keep the \textit{blackboxness} of BEAT. $f_{\theta'}$ can be an arbitrary network. \cref{eq:bayesianClassifier} then becomes:
\begin{align}
\label{eq:ptBayesianClassifier}
    p(y' | \mathbf{x}', \mathbf{x}, \Tilde{\mathbf{x}}, y) = E_{\theta'\sim p(\theta')}[p(y' | \mathbf{x}', \mathbf{x}, \Tilde{\mathbf{x}}, y, \theta, \theta')] \nonumber \\
    \approx \frac{1}{N}\sum_{i = 1}^Np(y'|\mathbf{x}',\theta'_i,\theta), \theta' \sim p(\theta' | \mathbf{x}, \Tilde{\mathbf{x}}, y, \theta)
\end{align}
We assume $\theta$ is obtained through pre-training. Then BEAT training can be conducted through alternative sampling:
\begin{align}
\label{eq:ptModelSampling}
    &\{\mathbf{x}, \Tilde{\mathbf{x}}, y\}_t | \theta, \theta'_{t-1} \sim p(\mathbf{x}, \Tilde{\mathbf{x}}, y | \theta, \theta'_{t-1}) \nonumber \\
    &\theta'_t | \{\mathbf{x}, \Tilde{\mathbf{x}}, y\}_t, \theta \sim p(\theta' | \{\mathbf{x}, \Tilde{\mathbf{x}}, y\}_t, \theta)
\end{align}
Since $f$ can be much smaller than $g$, BEAT on $f$ is faster than solving \cref{eq:logLikelihood} on $g$.  Following~\cref{eq:ptModelSampling}, the inference is conducted via alternatively solving \cref{eq:logLikelihood} and sampling $\theta'$. The mathematical derivations and algorithms for inference, with implementation details, are in Appendix B.

\section{Experiments}
\subsection{Experimental Settings}
\label{sec:setting}
We briefly introduce the experimental settings here, and the additional details are in Appendix D.

\paragraph{Datasets and Classifiers:} We choose three widely adopted benchmark datasets in HAR: HDM05~\cite{muller_documentation_2007}, NTU60~\cite{shahroudy_ntu_2016} and NTU120~\cite{liu_ntu_2020}. For base classifiers, we employ four recent classifiers: ST-GCN~\cite{yan_spatial_2018}, CTR-GCN~\cite{chen_channel-wise_2021}, SGN~\cite{zhang_semantics-guided_2020} and MS-G3D~\cite{liu_disentangling_2020}. Since the classifiers do not have the same setting (e.g. data needing sub-sampling~\cite{zhang_semantics-guided_2020}), we unify the data format. For NTU60 and NTU120, we sub-sample the frames to 60. For HDM05, we use a sliding window to divide the data into 60-frame samples~\cite{wang_understanding_2021}. Finally, we retrain the classifiers following their original settings.

\paragraph{Attack Setting:} After training the classifiers, we collect the correctly classified testing samples for attack. We employ state-of-the-art attackers designed for skeleton-based HAR: SMART ($l_2$ attack)~\cite{wang_understanding_2021}, CIASA ($l_{\infty}$ attack)~\cite{liu_adversarial_2020} and BASAR ($l_2$ attack)~\cite{diao_basarblack-box_2021}, and follow their default settings. Further, we use the untargeted attack, which is the most aggressive setting. Since all attackers are iterative approaches and more iterations lead to more aggressive attacks, we use 1000-iteration SMART (SMART-1000) on all datasets, 1000-iteration CIASA (CIASA-1000) on HDM05 and 100-iteration CIASA (CIASA-100) on NTU 60/120, since CIASA-1000 on NTU 60/120 is prohibitively slow (approximately 2 months on a Nvidia Titan GPU). We use the same iterations for BASAR as in their paper. Please refer to \textbf{Appendix D} for fuller evaluation details.

\paragraph{Defense Setting:} To our best knowledge, BEAT is the first black-box defense for skeleton-based HAR. So there is no method for direct comparison. There is a technical report~\cite{zheng_towards_2020} which is a simple direct application of randomized smoothing (RS)~\cite{Cohen19_ICML}. We use it as one baseline. Standard AT~\cite{madry_towards_2018} has recently been briefly attempted on HAR~\cite{tanaka2021adversarial}, so we use it as a baseline SMART-AT~\cite{BASAR_journal} which employs SMART as the attacker. We also employ another two baseline methods TRADES~\cite{TRADES} and MART~\cite{MART}, which are the state-of-the-art defense methods on images. We employ perturbations budget $\epsilon = 0.005$ for AT methods~\cite{madry_towards_2018,MART,TRADES} and compare other $\epsilon$ settings in \cref{sec:analysis}.

\paragraph{Computational Complexity:}  We use 20-iteration attack for training SMART-AT, TRADES and MART, since more iterations incur much higher computational overhead than BEAT, leading to unfair comparison. We compare the training time of BEAT with other defenses on all datasets in Appendix A. Since BEAT does not need to re-train the target model, it reduces training time (by 12.5\%-70\%) compared with the baselines. 

\paragraph{Appended Models:} Although $f_{\theta'}$ can be any model, a simple two-layer fully-connected layer network (with the same dimension as the original output) proves to work well in all cases. During attack, we attack the full model (with $f_{\theta'}$). We use five appended models in all experiments and explain the reason in the ablation study later.
\subsection{Attack Success and Adversary Quality}
\label{sec:evaluation}
\begin{table*}[ht]
    \centering
    \setlength{\tabcolsep}{1 mm}{
\resizebox{1\linewidth}{!}{
    \begin{tabular}{c| c c c|c c c|c c c|c c c}
        \hline
        \multirow{2}{*}{\small{HDM05}} & \multicolumn{3}{c| }{ST-GCN} & \multicolumn{3}{c| }{CTR-GCN} & \multicolumn{3}{c| }{SGN} & \multicolumn{3}{c }{MS-G3D}\\
         & \small{Accuracy}$\uparrow$ & \small{SMART}$\downarrow$ & \small{CIASA}$\downarrow$ & \small{Accuracy}$\uparrow$ & \small{SMART}$\downarrow$ & \small{CIASA}$\downarrow$ & \small{Accuracy}$\uparrow$ & \small{SMART}$\downarrow$ & \small{CIASA}$\downarrow$ & \small{Accuracy}$\uparrow$ & \small{SMART} $\downarrow$& \small{CIASA}$\downarrow$\\
         \hline
         \small{ST}& 93.22\% & 100\% & 100\% & 94.16\% & 89.51\% & 90.63\% & 94.16\% & 97.99\% & 99.60\% & 93.78\% & 96.80\% & 95.83\%\\
         \hline
         \small{SMART-AT}& 91.90\% & 88.62\% & 90.63\% & 93.03\% & 75.67\% & 80.31\% & 93.32\% & 96.65\% & 97.32\% & 92.84\% & 69.40\% & 84.77\%\\
         \hline
         \small{RS} & 92.66\% & 96.10\% & 96.88\% & 92.09\% & 81.47\% & 80.08\% & 92.81\% & 91.52\% & 98.43\% & 93.03\% & 94.92\% & 94.58\%\\
         \hline
         \small{MART} & 91.14\% & 80.80\% & 83.54\% & 91.53\% & 67.63\% & 77.73\% & 93.78\% & 97.10\% & 98.39\% & 91.52\% & 56.40\% & 80.13\%\\
         \hline
         \small{TRADES} & 91.53\% & 79.46\% & 85.00\% & 92.84\% & 73.88\% & 75.00\% & 92.28\% & 96.35\% & 99.91\% & 90.02\% & 55.74\% & 54.17\%\\
         \hline
         \small{BEAT(Ours)} & \textbf{93.03\%} & \textbf{61.46\%} & \textbf{62.50\%} & \textbf{93.22\%} & \textbf{64.91\%} & \textbf{65.96\%} & \textbf{94.72\%} & \textbf{26.82\%} & \textbf{27.60\%} & \textbf{93.60\%} & \textbf{20.20\%} & \textbf{20.46\%}\\
         \hline
         \hline
         \small{NTU60} & \small{Accuracy}$\uparrow$ & \small{SMART}$\downarrow$ & \small{CIASA}$\downarrow$ & \small{Accuracy}$\uparrow$ & \small{SMART}$\downarrow$ & \small{CIASA}$\downarrow$ & \small{Accuracy}$\uparrow$ & \small{SMART}$\downarrow$ & \small{CIASA}$\downarrow$ & \small{Accuracy}$\uparrow$ & \small{SMART}$\downarrow$ & \small{CIASA}$\downarrow$\\
         \hline
         \small{ST}& 76.81\% & 100\% & 99.36\% & 82.90\% & 100\% & 100\% & 86.22\% & 100\% & 97.36\% & 89.36\% & 99.67\% & 97.72\% \\
         \hline
         \small{SMART-AT} & 72.80\% & 100\% & 85.47\% & \textbf{83.70\%} & 99.99\% & 76.54\% & 83.26\% & 100\% & 87.34\% & 87.79\% & 100\% & 53.96\%\\
         \hline
         \small{RS} & 75.89\% & 100\% & 94.51\% & 82.67\% & 100\% & 91.95\% & 83.02\% & 100\% & 92.85\% & 88.12\% & 100\% & 88.70\% \\
         \hline
         \small{MART} & 71.94\% & 100\% & 79.68\% & 80.31\% & 100\% & 79.63\% & 83.21\% & 99.95\% & 83.13\% & 85.35\% & 100\% & 50.61\%\\
         \hline
         \small{TRADES} & 71.40\% & 100\% & 73.75\% & 79.63\% & 100\% & 76.90\% & 82.30\% & 100\% & 76.56\% & 85.16\% & 99.96\% & 48.52\%\\
         \hline
         \small{BEAT(Ours)} & \textbf{76.45\%} & \textbf{62.99\%} & \textbf{71.28\%} & 82.77\% & \textbf{73.16\%} & \textbf{62.85\%} & \textbf{86.07\%} & \textbf{40.10\%} & \textbf{34.31\%} & \textbf{88.78\%} & \textbf{32.31\%} & \textbf{34.21\%}\\
         \hline
         \hline
         \small{NTU120} & \small{Accuracy}$\uparrow$ & \small{SMART}$\downarrow$ & \small{CIASA}$\downarrow$ & \small{Accuracy}$\uparrow$ & \small{SMART}$\downarrow$ & \small{CIASA}$\downarrow$ & \small{Accuracy}$\uparrow$ & \small{SMART}$\downarrow$ & \small{CIASA}$\downarrow$ & \small{Accuracy}$\uparrow$ & \small{SMART}$\downarrow$ & \small{CIASA}$\downarrow$\\
         \hline
         \small{ST}& 68.34\% & 100\% & 99.20\% & 74.59\% & 99.80\% & 99.44\% &74.15\% &99.94\% &99.20\% &84.71\% &99.47\% &97.81\%\\
         \hline
         \small{SMART-AT} & 67.28\% & 100\% & 84.90\% & \textbf{75.89\%} & 100\% & 83.24\% & 71.30\% & 100\% & 94.63\% & 81.90\% & 99.40\% & 63.50\% \\
         \hline
         \small{RS} & 66.81\% & 100\% & 95.55\% & 74.04\% & 100\% & 95.15\% & 71.40\% & 100\% & 98.83\% & 82.17\% & 99.93\% & 98.40\%\\
         \hline
         \small{MART} & 58.43\% & 99.83\% & 80.69\% & 70.54\% & 99.90\% & 80.46\% & 70.05\% & 99.84\% & 86.04\% & 78.89\% & 99.88\% & 57.64\%\\
         \hline
         \small{TRADES} & 61.41\% & 99.67\% & 82.80\% & 71.99\% & 100\% & 81.32\% & 69.37\% & 99.95\% & 83.72\% & 79.04\% & 99.95\% & 55.36\%\\
         \hline
         \small{BEAT(Ours)} & \textbf{68.34\%} & \textbf{84.55\%} & \textbf{77.19\%} & 74.59\% & \textbf{85.66\%} & \textbf{77.82\%} & \textbf{73.53\%} & \textbf{56.44\%} & \textbf{37.22\%} & \textbf{84.70\%} & \textbf{41.61\%} & \textbf{39.87\%}\\
         \hline

    \end{tabular}}
}
    \caption{The results of BEAT and other 5 methods. ST means standard training. ‘SMART’ and ‘CIASA’ are the attack success rate of SMART and CIASA. We show the best performance with bold in all 5 defense methods (not include ST).}
    \label{tab:comparisons}
\end{table*}
\paragraph{Robustness under White-box Attacks} We show the results of all models on all datasets in \cref{tab:comparisons}. First, BEAT does not severely compromise standard accuracy across models and data. The BEAT accuracy is within a small range (+0.6/-0.9\%) from that of the standard training, in contrast to the often noticeable accuracy decreased in other defense methods. Next, BEAT has the best robustness in all 144 scenarios (training methods vs. classifiers vs. datasets vs. attackers) and 
often by big margins, especially under extreme SMART-1000 and CIASA-1000 attacks. Overall, BEAT can significantly improve the adversarial robustness and eliminate the accuracy-robustness trade-off.

\paragraph{Robustness under Black-box Attacks}
Black-box attack in skeletal HAR is either transfer-based~\cite{wang_understanding_2021} or decision-based~\cite{diao_basarblack-box_2021}. However, existing white-box attacks (SMART and CIASA) are highly sensitive to the chosen surrogate and the target classifier. According to our preliminary experiments (see Appendix A), transfer-based SMART often fails when certain models are chosen as the surrogate, which suggests that transfer-based attack is not a reliable way of evaluating defense in skeletal HAR. Therefore we do not employ transfer-based attack for evaluation. BASAR is a decision-based approach, which is truly black-box and has proven to be far more aggressive and shrewd~\cite{diao_basarblack-box_2021}. We employ its evaluation metrics, i.e. the averaged $l_2$ joint position deviation ($l$), averaged $l_2$ joint acceleration deviation ($\Delta a$) and averaged bone length violation percentage ($\Delta B/B$), which all highly correlate to the attack imperceptibility. We randomly select samples following~\cite{diao_basarblack-box_2021} for attack. The results are shown in~\cref{tab:basar}. BEAT can often reduce the quality of adversarial samples, which is reflected in $l_2$, $\Delta a$ and $\Delta B/B$. The increase in these metrics means severer jittering/larger deviations from the original motions, which is very visible and raises suspicion. We show one example in~\cref{fig:BASAR} and fuller results are in Appendix A.
\begin{figure}[!tb]
    \centering
    \includegraphics[width=\linewidth]{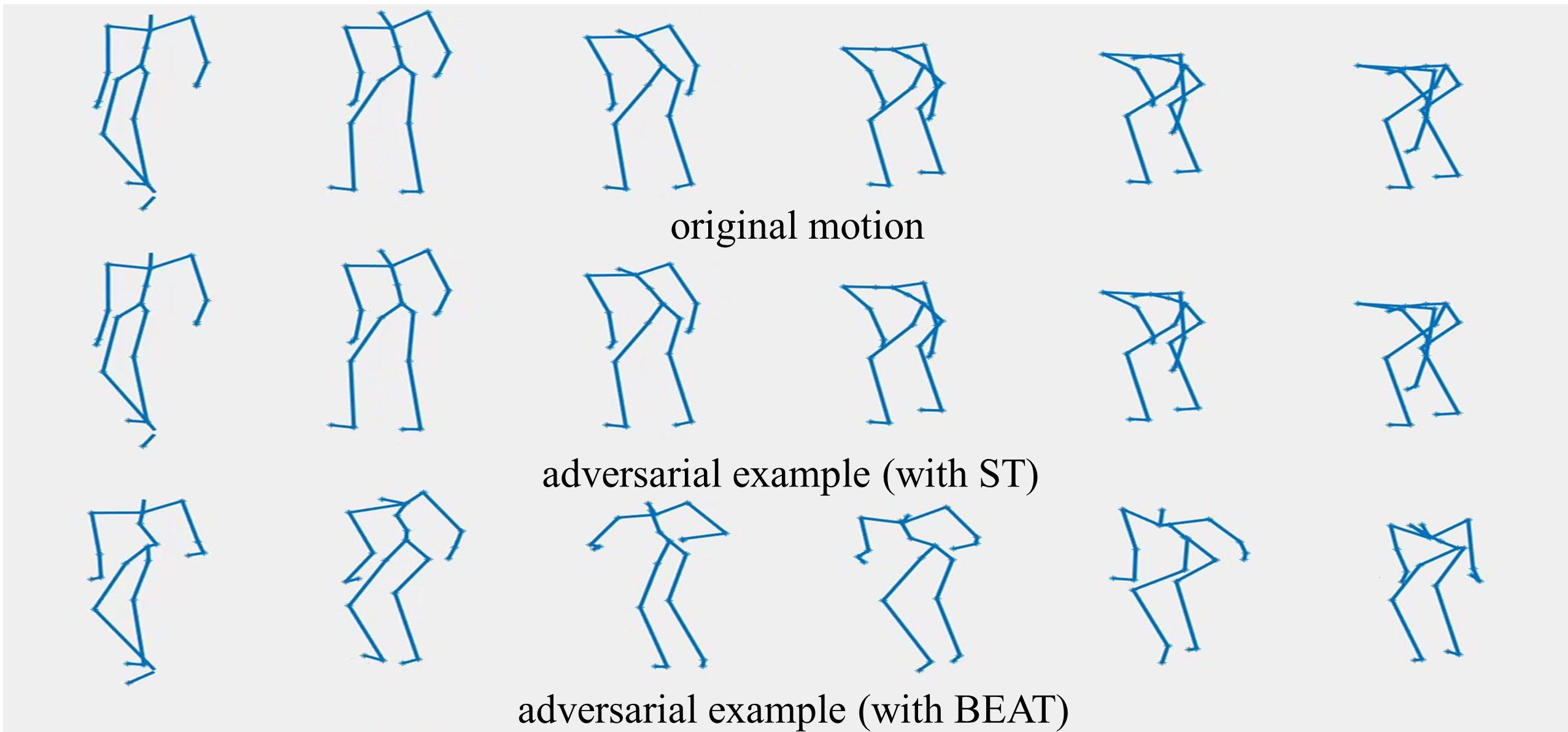}
    \caption{The original motion `deposit' (top) is attacked by BASAR on standard trained model (middle) and BEAT trained model (bottom) separately. The visual difference of the original motion and the attacked motion on BEAT is bigger, compared with the attacked motion on ST.}
    \label{fig:BASAR}
\end{figure}

\begin{table}[htb]
\centering
\setlength{\tabcolsep}{1mm}{
\resizebox{1.0\linewidth}{!}{
\begin{tabular}{c|c|c|c|c}
\hline
& \small{STGCN} & \small{CTRGCN} & \small{SGN} & \small{MSG3D} \\
\hline
$l$ $\uparrow$ & 0.77/\textbf{0.82} & 0.67/\textbf{0.79} & 0.84/\textbf{1.05} & 0.20/\textbf{0.28} \\
\hline
$\Delta a$ $\uparrow$ & 0.21/\textbf{0.22} & 0.14/0.14 & 0.05/\textbf{0.07} & 0.086/\textbf{0.095} \\
\hline
$\Delta B/B$ $\uparrow$ & 0.42\%/\textbf{0.77\%} & 0.80\%/\textbf{0.94\%} & 1.1\%/\textbf{1.5\%} & 1.1\%/\textbf{1.2\%} \\
\hline
\hline
$l$ $\uparrow$ & 0.03/\textbf{0.05} & 0.05/\textbf{0.06} & 0.06/\textbf{0.08} & 0.09/0.09 \\
\hline
$\Delta a$ $\uparrow$ & 0.015/\textbf{0.017} & 0.02/\textbf{0.03} & 0.003/\textbf{0.004} & 0.03/\textbf{0.04} \\
\hline
$\Delta B/B$ $\uparrow$ & 4.2\%/\textbf{4.8\%} & 6.5\%/\textbf{7.4\%} & 1.3\%/\textbf{1.7\%} & 8.9\%/\textbf{11.0\%} \\
\hline
\hline
$l$ $\uparrow$ & 0.03/\textbf{0.04} & 0.04/\textbf{0.06} & 0.087/\textbf{0.103} & 0.06/\textbf{0.08} \\
\hline
$\Delta a$ $\uparrow$ & 0.015/\textbf{0.018} & 0.019/\textbf{0.022} & 0.005/\textbf{0.006} & 0.02/\textbf{0.03} \\
\hline
$\Delta B/B$ $\uparrow$ & 4.0\%/\textbf{4.7\%} & 5.4\%/\textbf{5.6\%} & 2.3\%/\textbf{2.7\%} & 6.8\%/\textbf{9.0\%} \\
\hline
\end{tabular}}}
\caption{Untargeted attack on HDM05 (top), NTU60 (middle) and NTU120 (bottom) from BASAR. xxx/xxx is pre/post BEAT results.}
\label{tab:basar}
\end{table}

\subsection{Additional Performance Analysis}
\label{sec:analysis}
\begin{figure}[!tb]
    \centering
    \includegraphics[width=\linewidth]{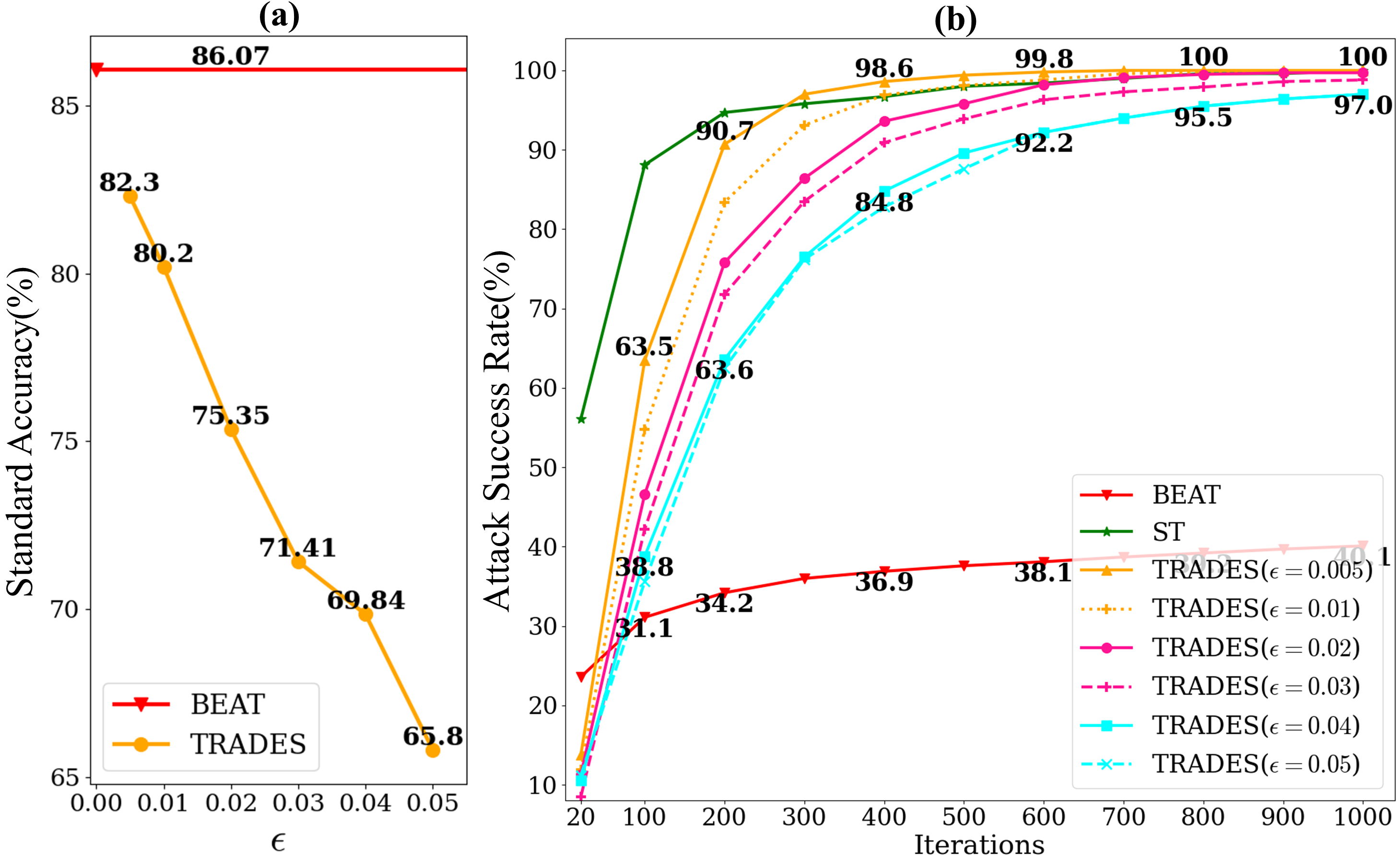}
    \caption{Comparisons with TRADES with different perturbation budget ($\epsilon$) on NTU60 with SGN. (a): standard accuracy vs. $\epsilon$; (b): results against SMART with 20 to 1000 iterations.}
    \label{fig:acc_epsilon}
\end{figure}
\paragraph{Comparison with other AT methods.} As shown in ~\cref{tab:comparisons}, all baseline methods perform worse than BEAT, sometimes completely fail, e.g. failing to defend against SMART-1000 in large-scale datasets (NTU 60 and NTU 120). After investigating their defenses against SMART from iteration 20 to 1000 (Appendix A), we found the key reason is the baseline methods overly rely on the aggressiveness of the adversaries sampled during training. To verify it, we increase the perturbation budget $\epsilon$ from 0.005 to 0.05 during training in TRADES, and plot their standard accuracy \& attack success rate (ASR) vs. $\epsilon$ in~\cref{fig:acc_epsilon}. Note that BEAT does not rely on a specific attacker. We find TRADES is highly sensitive to $\epsilon$ values: larger perturbations in adversarial training improve the defense (albeit still less effective than BEAT), but harm the standard accuracy (\cref{fig:acc_epsilon}(a)). Further, sampling adversaries with more iterations (e.g. 1000 iterations) during AT may also improve the robustness (still worse than BEAT~\cref{fig:acc_epsilon}(b)) but is prohibitively slow, while BEAT requires much smaller computational overhead (see \cref{sec:setting}).


\paragraph{Does BEAT Obfuscate Gradients?}
Since BEAT averages the gradient across different models, we investigate whether its robustness is due to obfuscated gradients, because obfuscated gradients can be circumvented and are not truly robust~\cite{athalye2018obfuscated}. One way to verify this is to test BEAT on black-box attacks~\cite{athalye2018obfuscated}, which we have demonstrated in \cref{sec:evaluation}. To be certain, we design another attack which can also circumvent defense relying on obfuscated gradients. We deploy an adaptive attack called EoT-SMART based on~\cite{tramer2020adaptive}: in each step, we estimate the expected gradient by averaging the gradients of multiple randomly interpolated samples. \cref{tab:EOT} shows that EoT-SMART performs only slightly better than SMART, demonstrating that BEAT does not rely on obfuscated gradients.   

\begin{table}[htb]
\centering
\setlength{\tabcolsep}{0.5 mm}{
\resizebox{1.0\linewidth}{!}{
\begin{tabular}{c|c|c|c|c}
\hline
BEAT & ST-GCN & CTR-GCN  &SGN  &MS-G3D  \\ \hline
HDM05 & 62.29\% (+0.8\%) &68.40\% (+3.5\%)  &27.31\% (+0.5\%)  & 21.43\% (+1.2\%) \\ \hline
NTU 60 & 63.23\% (+0.2\%) &73.03\% (-0.1\%)  &41.99\% (+1.9\%)  & 32.30\% (-0.0\%) \\ \hline
\end{tabular}}}
\caption{Attack success rate (ASR) of EOT-SMART. ($\pm$xx\%) means the ASR difference with SMART.}
\label{tab:EOT}
\end{table}




\subsection{Ablation Study}
\paragraph{Number of Appended Models} Although BNNs theoretically require sampling of many models for inference, in practice, we find a small number of models suffice. To show this, we conduct an ablation study on the number of appended models (N in \cref{eq:ptBayesianClassifier}). As shown in \cref{tab:ablation}, with $N$ increasing, BEAT significantly lowers the attack success rates, which shows the Bayesian treatment of the model parameters is able to greatly increase the robustness. Further, when N $>$ 5, there is a diminishing gain with a slight improvement in robustness but also with increased computation. So we use N=5 by default. 

\begin{figure}[!t]
    \centering
    \includegraphics[width=\linewidth]{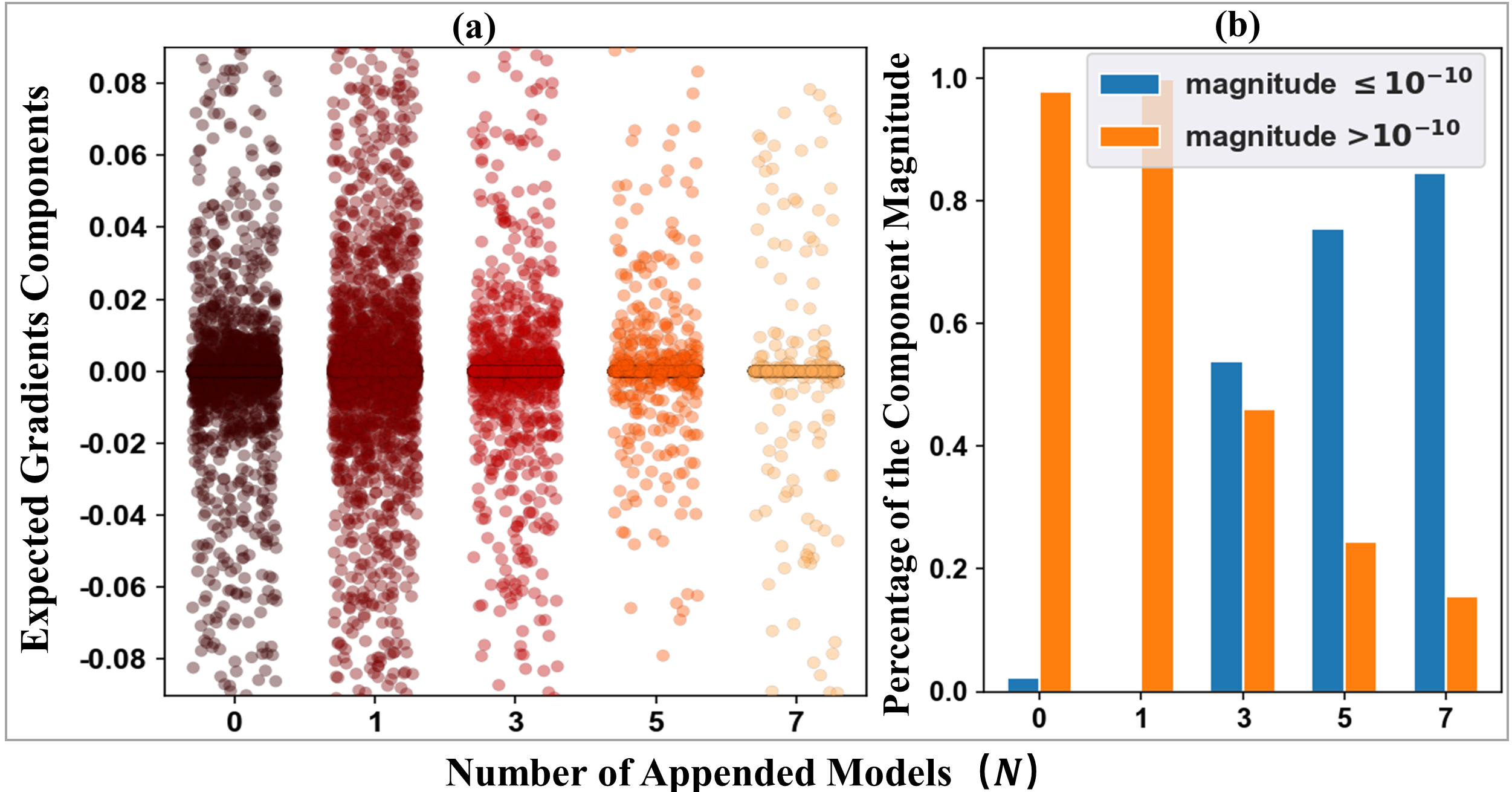}
    \caption{The components of the expected loss gradients of BEAT on NTU60 with SGN. $N=0$ is standard training. (a): the empirical distribution of gradient components; (b): the percentage of the expected gradient components with magnitude above and below 10$^{-10}$.}
    \label{fig:loss_gradient}
\end{figure}

We further show why our \textit{post-train} Bayesian strategy is able to greatly increase the robustness. The classification loss gradient with respect to data is key to many attack methods. In a deterministic model, this gradient is computed on one model; in BEAT, this gradient is averaged over all models, i.e. the expected loss gradient. Theoretically, with an infinitely wide network in the large data limit, the expected loss gradient achieves 0, which is the source of the good robustness of BNNs~\cite{Bortolussi_onTheRobustness_2022}. To investigate whether BEAT's robustness benefits from the vanishing expected gradient, we randomly sample 500 motions from the testing data and sample one frame from each motion. Then we compute their expected loss gradients and plot a total of 37500 loss gradient components in~\cref{fig:loss_gradient} (a), where each dot represents a component of the expected loss gradient of one frame. \cref{fig:loss_gradient} (b) shows the percentage of the expected gradient components close to 0. \Cref{fig:loss_gradient} essentially shows the empirical distribution of the component-wise expected loss gradient. When $N$ increases, the gradient components steadily approach zero, indicating a vanishing expected loss gradient which provides robustness~\cite{Bortolussi_onTheRobustness_2022}.

\paragraph{Joint Distribution of Data and Adversaries} Other than the Bayesian treatment of models, BEAT also benefits from the Bayesian treatment on the adversaries. To see its contribution, we plug-and-play our post-train Bayesian strategy to other AT methods which do not model the adversarial distribution. Specifically, we design a post-train Bayesian TRADES (PB+TRADES) with different number of appended models, and compare them with BEAT in~\cref{tab:ablation}. While both BEAT and PB+TRADES benefit from BNNs, BEAT still outperforms PB+TRADES by large margins. Note the major difference between BEAT and PB+TRADES is whether to consider the full adversarial distribution, which shows the benefit of bringing the full adversarial distribution into the joint probability.
\begin{table}[t]
    \centering
    \resizebox{0.9\linewidth}{!}{
    \begin{tabular}{c|c c|c c}
        \hline
         \multirow{2}{*}{N} & \multicolumn{2}{c|}{Standard Accuracy$\uparrow$} & \multicolumn{2}{c}{Attack Success Rate$\downarrow$} \\
          & BEAT &PB+TRADES & BEAT &PB+TRADES \\
         \hline
         1 & 84.90& 83.66\% & 96.39\% & 97.40\%\\
         \hline
        3 & 85.72\%& 84.96\% & 57.13\% & 80.37\%\\
         \hline
        5 & 86.07\%& 84.74\% & 40.10\% & 59.03\%\\
        \hline
        7 & 86.01\%& 84.86\% & 27.40\% & 48.97\%\\
         \hline
    \end{tabular}}
        \caption{Ablation Study on NTU 60 with SGN as the base classifier. `N'is the number of appended models. `PB' means Post-train Bayesian strategy. `Attack Success Rate' is the defense results against SMART-1000.}
    \label{tab:ablation}
\end{table} 
\section{Discussion, Conclusions and Future Work}
One limitation is prior knowledge is needed on $d(\mathbf{x}, \Tilde{\mathbf{x}})$ in \cref{eq:fullJointEnergy}, either explicitly as BEAT or implicitly e.g. using another model to learn the data manifold. However, this is lightweight as manifold learning/representation is a rather active field and many methods could be used. BEAT can potentially incorporate any manifold representation. Also, we assume that all adversarial samples are distributed closely to the data manifold, which is true for images~\cite{stutz2019disentangling} and skeletal motion~\cite{diao_basarblack-box_2021}, but not necessarily for other data.

To our best knowledge, we proposed the first black-box defense for skeleton-based HAR. Our method BEAT is underpinned by a new Bayesian Energy-based Adversarial Training framework, and is evaluated across various classifiers, datasets and attackers. Our method employs a post-train strategy for fast training and a full Bayesian treatment on normal data, the adversarial samples and the classifier, without adding much extra computational cost. In future, we will extend BEAT to more data types, both time-series and static, such as videos and stock prices, as well as implicit manifold parameterization for images, by employing task/data specific $d(\mathbf{x}, \Tilde{\mathbf{x}})$ in \cref{eq:adDist}.

\section{Acknowledgment}
This project has received funding from the European Union’s Horizon 2020 research and innovation programme under grant agreement No 899739 CrowdDNA.

\bibliography{aaai23}

\end{document}


\maketitle


\appendix 

\section{More Results and Analysis}
\begin{figure*}[!tb]
    \centering
    \includegraphics[width=\linewidth]{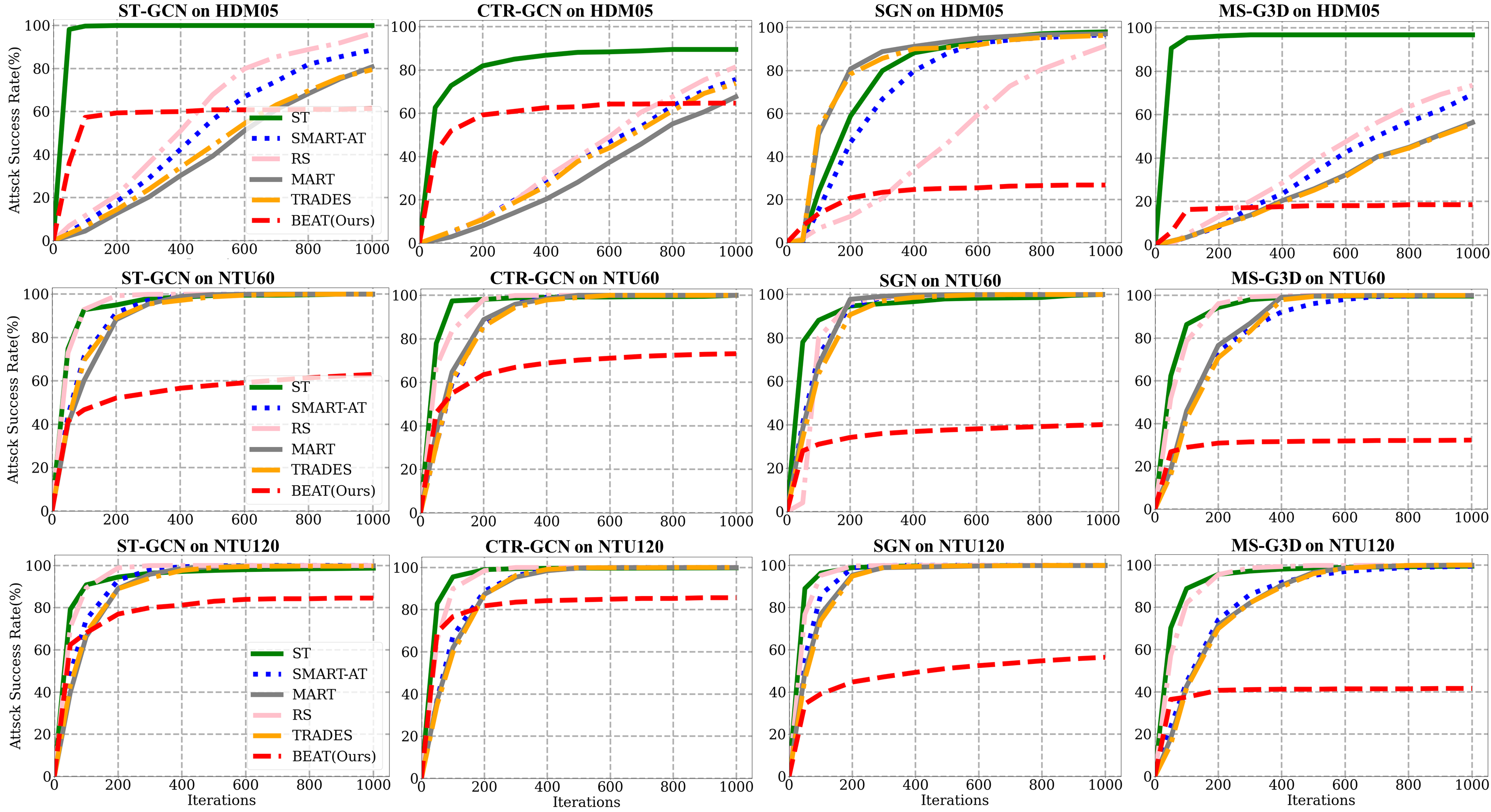}
    \caption{The attack success rate vs. attack strength curves against SMART. For each subplot, the abscissa axis is iterations while the ordinate axis is the attack success rate(\%). ST means standard training.}
    \label{fig:attack_strength}
\end{figure*}

\paragraph{Computational complexity.} 
We used one RTX 3090 GPU for all experiments, except for training BEAT for MS-G3D on NTU60/120 with two Tesla V100 (32G) GPUs. All experiments are conducted on the Pytorch platform. The training time on different combinations of models and datasets are shown in \cref{tab:time}. Since BEAT does not need to re-train the target model, it is the least time-consuming among all AT methods~\cite{madry_towards_2018,TRADES,MART}, e.g. only requiring 12.5\%-70\% of the training time of others. Although RS needs less training time than BEAT, its overall robustness is low and therefore fails to defend against attacks, as shown in \cref{fig:attack_strength}.   
\begin{table}[tb]
\centering
\resizebox{1.0\linewidth}{!}{
\begin{tabular}{c|ccccc}
\hline
HDM05   & SMART-AT & RS & MART & TRADES & BEAT(Ours) \\ \hline
ST-GCN  & 3.6   & 0.3   &  3.6    &  5.1      & 2.2     \\ 
CTR-GCN & 7.6   & 0.7   & 7.7     &  10.1      & 1.3     \\ 
SGN     & 1.4   & 0.1   &  1.4    &  1.1      & 0.7     \\ 
MS-G3D  & 4.7   & 0.5   &   4.7   &   6.6     & 3.2     \\ \hline \hline
NTU 60   & SMART-AT & RS & MART & TRADES & BEAT(Ours) \\ \hline
ST-GCN  & 20   & 6   &  21    &  29      & 8     \\ 
CTR-GCN & 67   & 6   &  68    &  96      & 24     \\ 
SGN     & 11   & 2   &  11    &  12      & 6.8     \\ 
MS-G3D  & 103   & 24   &  103    &   170     & 72     \\ \hline \hline
NTU 120   & SMART-AT & RS & MART & TRADES & BEAT(Ours) \\ \hline
ST-GCN  & 31   & 2   &   31   &  43      & 10     \\ 
CTR-GCN & 52   & 3   &  52    &   68     & 13     \\ 
SGN     & 18   & 5   &  18    &   20     & 5     \\ 
MS-G3D  & 180   & 45   &  180    &  288      & 115     \\ \hline 
\end{tabular}}
\caption{Training time(hours) of all defense methods.}
\label{tab:time}
\end{table}

\paragraph{Further ablation study.}
\textit{Post-train} BEAT keeps the pre-trained classifier intact and append a tiny model with parameters $\theta'$ behind the classifier using a skip connection: logits = $f_{\theta'}(\phi(\mathbf{x})) + g_\theta(\mathbf{x})$, in contrast to the original logits = $g_\theta(\mathbf{x})$. $\phi(\mathbf{x})$ can be the latent features of $\mathbf{x}$ or the original logits $\phi(\mathbf{x}) = g_\theta(\mathbf{x})$. We conduct an ablation study on STGCN and HDM05 to demonstrate the effectiveness of the both settings. Shown in \cref{tab:ablation2}, the two settings achieved similar results. However, the model choice has implications. When $\phi(\mathbf{x}) = g_\theta(\mathbf{x})$ is employed, it can help treating $g$ as a black-box classifier, requiring little prior knowledge about the classifier itself. Arguably, this is a better setting, where the people doing BEAT can largely ignore the classifier details as they might not be the people who developed the classifier. So we employ $\phi(\mathbf{x}) = g_\theta(\mathbf{x})$ to keep the black-boxness of BEAT.

\begin{table}[tb]
    \centering
    \resizebox{1.0\linewidth}{!}{
    \begin{tabular}{c|c|c|c}
        \hline
         $\phi(\mathbf{x})$ & Accuracy$\uparrow$ & SMART$\downarrow$ & CIASA$\downarrow$ \\
         \hline
         Latent Features of $\mathbf{x}$ & 92.84\% & 61.61\% & 64.68\%\\
         \hline
         Original Logits $g_\theta(\mathbf{x})$ & 93.03\% & 61.46\% & 62.50\%\\
         \hline
    \end{tabular}}
        \caption{Ablation Study (various $\phi(\mathbf{x})$ settings) on HDM05 with STGCN as the base classifier. ‘SMART’ and ‘CIASA’ are the attack success rate of SMART-1000 and CIASA-1000.}
    \label{tab:ablation2}
\end{table}

\paragraph{Further results of white-box attack.} 
We plot the robustness vs. attack strength (iterations) curves under SMART attack in \cref{fig:attack_strength}, which shows the defense performance under different attack strengths. Albeit worse than BEAT, most baseline methods can defend against attacks to certain extent when the attack is weak, i.e. a small number of iterations. However, as the attack strength increases, their robustness drops rapidly till failing completely.

\paragraph{Further results of black-box attack.} 
Here we first report the defense evaluation under transfer-based SMART attack. As shown in \cref{tab:transfer}, when using 2S-AGCN~\cite{Shi_2019_CVPR} as the surrogate model, the attack success rate is low on three of the four targeted models. This confirms the sensitivity of SMART on the chosen surrogate during transfer-based attack~\cite{wang_understanding_2021} and suggests that transfer-based attack is not a reliable way of evaluating defense. Therefore, we do not employ it to test the robustness of BEAT. 
\begin{table}[!htb]
    \centering
    \setlength{\tabcolsep}{1 mm}{
\resizebox{1\linewidth}{!}{
\begin{tabular}{c|c|c|c|c|c}
\hline
NTU 60   &      & ST-GCN & CTR-GCN & SGN & MSG3D \\\hline
\multirow{2}{*}{\begin{tabular}[c]{@{}c@{}}SMART\\ \end{tabular}} & ST   &  1.56\%      & 0.94\%    & 74.10\%    &0.2\%       \\
 & BEAT &  0.60\%      & 0.63\%    & 70.12\%    &0.2\%      \\\hline
\end{tabular}}}
\caption{Defense against transfer-based SMART. `SMART' is the attack success rate of transfer-based SMART, using 2S-AGCN as the surrogate model. `ST' means standard training. The attack success rate for transfer-based SMART is low on ST-GCN, CTR-GCN and MSG3D in both ST and BEAT.}
\label{tab:transfer}
\end{table}

To test defense against black-box attack, we employ BASAR~\cite{diao_basarblack-box_2021}. As BASAR can always achieve 100\% attack success, we need to evaluate the quality of the adversarial samples. The lower the quality is, the more effective our defense is. We employ several metrics to measure the attack quality, following ~\cite{diao_basarblack-box_2021}. They are the $l_2$ joint position deviation ($l$), $l_2$ joint acceleration deviation ($\Delta a$), $l_2$ joint angular acceleration deviation ($\Delta\alpha$), bone length violation percentage ($\Delta B/B$) and on-manifold sample percentage (OM). To comprehensively evaluate the defense performance, we show the maximum and mean value of these metrics, and the percentage of $l\ge a_1$ and $\Delta B/B \ge a_2$, where $a_1$ and $a_2$ are the thresholds. The mean metric gives the general quality of the attack, the max shows the worst quality of the attack and the percentage shows how many adversarial samples are above pre-defined thresholds on different metrics. The thresholds are chosen empirically based on classifiers and datasets, as the visual quality of adversarial samples vary. We chose the thresholds beyond which the visual difference between adversarial samples and the original motions become noticeable.

\begin{figure}[!tb]
    \centering
    \includegraphics[width=\linewidth]{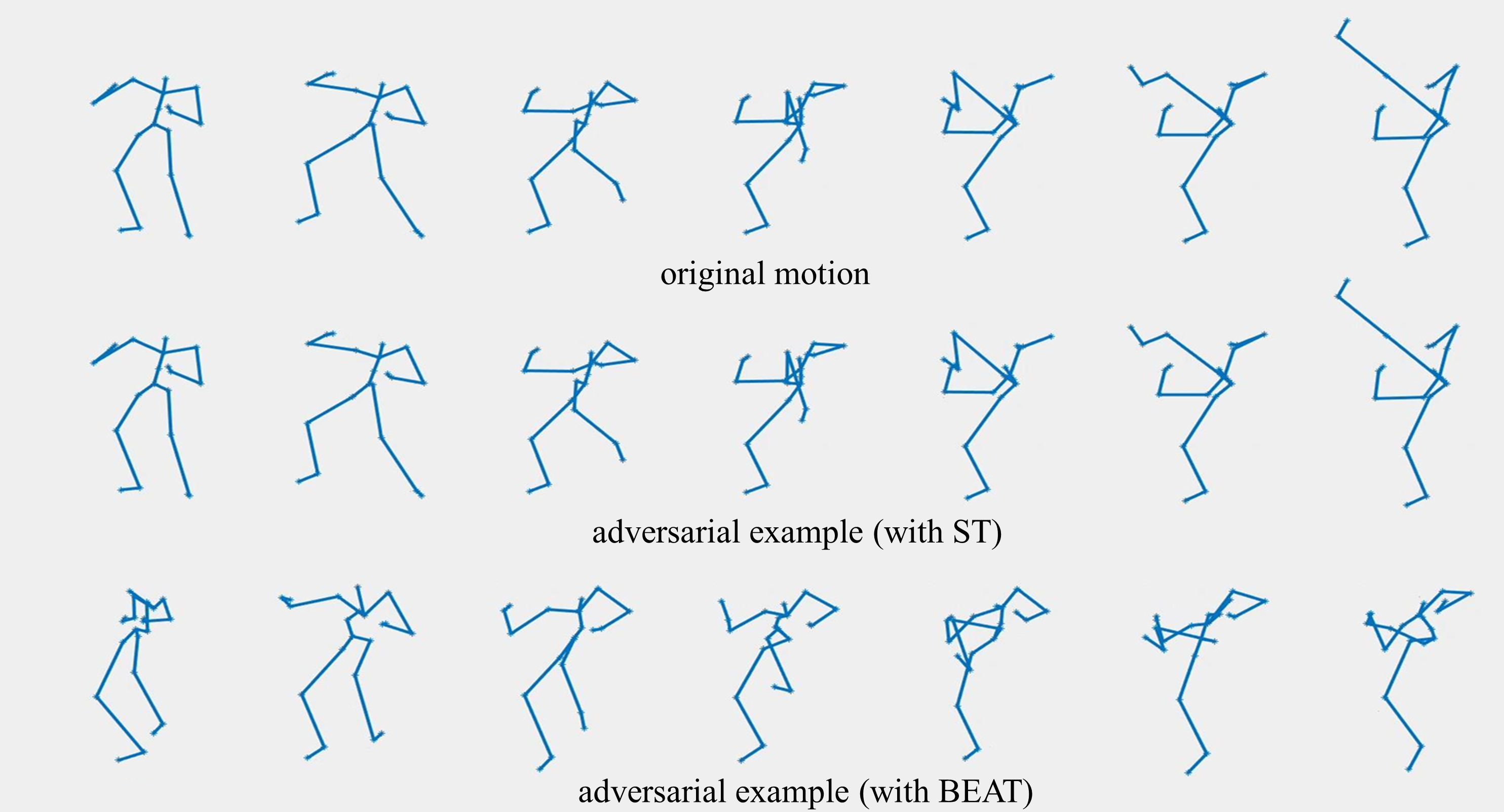}
    \caption{The original motion `kick' (top) is attacked by BASAR on standard trained model (middle) and BEAT trained model (bottom) separately. The adversarial sample under BEAT becomes visually different from the original motion.}
    \label{fig:BASAR1}
\end{figure}

\begin{figure}[!tb]
    \centering
    \includegraphics[width=\linewidth]{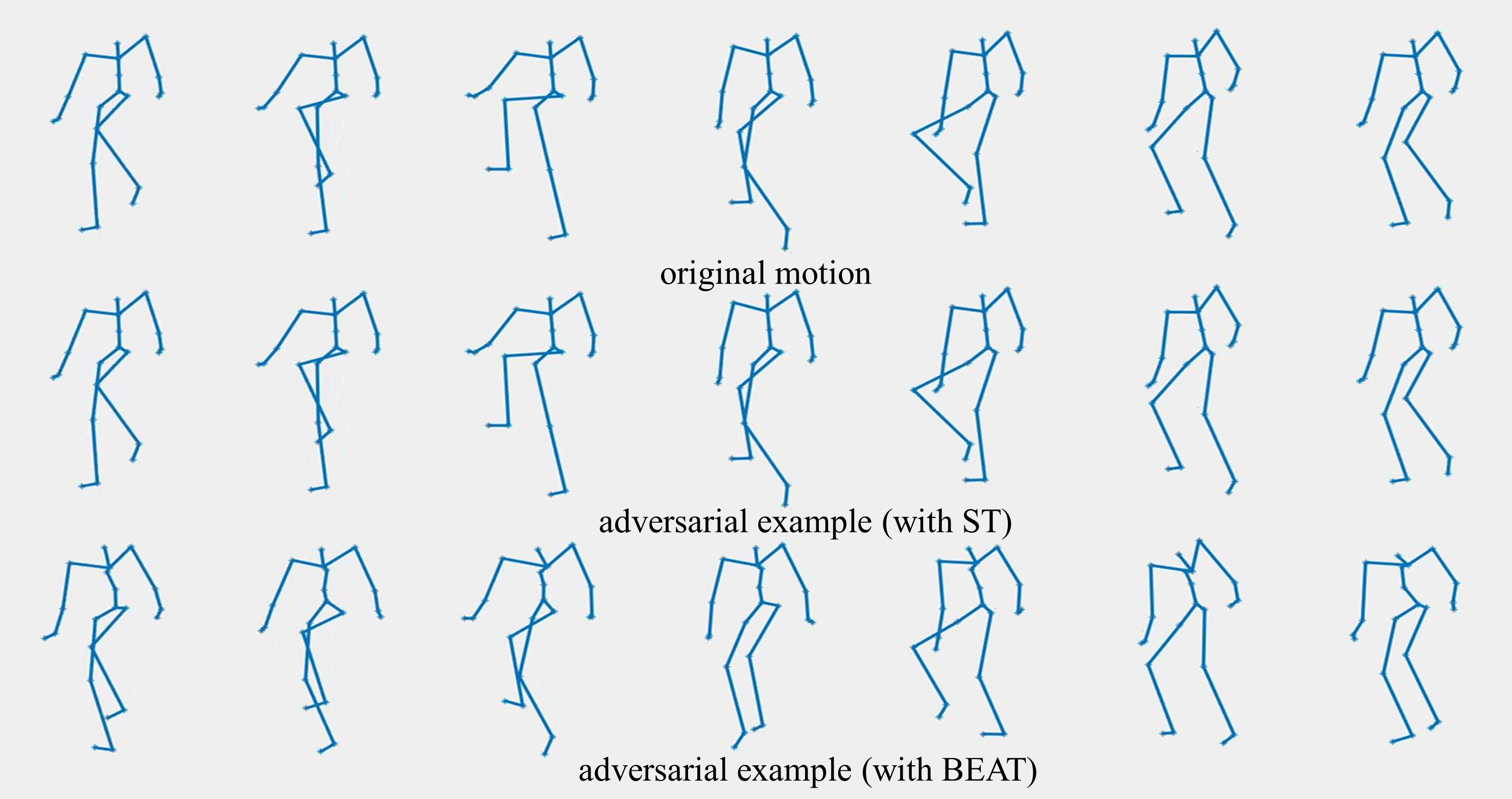}
    \caption{The original motion `climb' (top) is attacked by BASAR on standard trained model (middle) and BEAT trained model (bottom) separately. The adversarial sample under BEAT becomes visually different from the original motion.}
    \label{fig:BASAR2}
\end{figure}

The results are shown in \cref{tab:black_stgcn}-\ref{tab:black_msg3d}. First, BEAT can often reduce the quality of adversarial samples as shown in mean, max and thresholded metrics. This is reflected in $l_2$, $\Delta a$, $\Delta\alpha$ and  $\Delta B/B$. The increase in these metrics normally means there is severe jittering starting to appear in the adversarial samples, which is very visible and can raise suspicion. These attacks tend to fail in that humans would be able to detect them. Further, BEAT can often make adversarial samples more perceptible as OM decreases. It means more adversarial samples deviate from the normal data manifold, suggesting an overall lower quality of the adversarial samples, after BEAT training. Here we provide more visual results in \cref{fig:BASAR1} and \cref{fig:BASAR2}.

\begin{table*}[!tb]
    \centering
    \setlength{\tabcolsep}{1.2 mm}{
\resizebox{1.0\linewidth}{!}{
    \begin{tabular}{c|c|cc|cccc|ccccc}
    \hline
    STGCN &  &\multicolumn{2}{c|}{Threshold}  & \multicolumn{4}{c|}{Max} & \multicolumn{5}{c}{Mean} \\
    HDM05 &Accuracy& $l\ge3\uparrow$ & $\Delta B/B\ge5\%\uparrow$  & $l\uparrow$     & $\Delta a\uparrow$     & $\Delta\alpha\uparrow$    & $\Delta B/B\uparrow$    & $l\uparrow$   & $\Delta a~\uparrow$  & $\Delta\alpha\uparrow$  & $\Delta B/B~\uparrow$  & OM$\downarrow$  \\ \hline
    ST      &92.66\%   & 3.6\% &2.3\%   &4.2  &0.96 &1.52  &4.2\% & 0.77  &0.21 &0.41    &0.42\%    & 90.55\%   \\ \hline
    BEAT(Ours) &92.46\%& \textbf{6.4\%} &\textbf{5.4\%}   &\textbf{5.6}  &\textbf{1.32} & \textbf{2.91} &\textbf{5.3\%} & \textbf{0.82}    &\textbf{0.22} &\textbf{0.46}    &\textbf{0.77\%}    & \textbf{78.36\%}   \\ \hline\hline
    NTU60 &Accuracy& $l\ge0.1\uparrow$ & $\Delta B/B\ge10\%\uparrow$  & $l\uparrow$     & $\Delta a\uparrow$     & $\Delta\alpha\uparrow$    & $\Delta B/B\uparrow$    & $l\uparrow$   & $\Delta a\uparrow$  & $\Delta\alpha\uparrow$  & $\Delta B/B\uparrow$  & OM$\downarrow$  \\ \hline
    ST        &76.81\% & 3.7\% &8.8\%   &0.22  &0.08 & 0.13 &19.4\% & 0.03    &0.015 &0.023    &4.2\%    & 0\%   \\ \hline
    BEAT(Ours) &74.47\%& \textbf{8.1\%} &\textbf{14.4\%}   &\textbf{0.98}  &\textbf{0.11} & \textbf{0.16} &\textbf{43.0\%} & \textbf{0.05}    &\textbf{0.017} &\textbf{0.024}    &\textbf{4.8\%}    & 0\%   \\ \hline\hline
    NTU120 &Accuracy& $l\ge0.1\uparrow$ & $\Delta B/B\ge10\%\uparrow$  & $l\uparrow$     & $\Delta a\uparrow$     & $\Delta\alpha\uparrow$    & $\Delta B/B\uparrow$    & $l\uparrow$   & $\Delta a\uparrow$  & $\Delta\alpha\uparrow$  & $\Delta B/B\uparrow$  & OM$\downarrow$  \\ \hline
    ST         &68.34\%& 5.1\% &5.1\%   &0.19  &0.08 & 0.17 &29.3\% & 0.03    &0.015 &0.016    &4.0\%    & 3.8\%   \\ \hline
    BEAT(Ours) &66.84\%& \textbf{7.6\%} &\textbf{8.7\%}   &\textbf{0.32}  &\textbf{0.14} & \textbf{0.26} &\textbf{37.4\%} & \textbf{0.04}    &\textbf{0.018} &\textbf{0.02}    &\textbf{4.7\%}    & \textbf{3.3\%}   \\ \hline
    \end{tabular}}
}
    \caption{Defense results on STGCN. ST means standard training. Metrics are computed on the adversarial samples computed after ST training and BEAT training.}
    \label{tab:black_stgcn}
\end{table*}
\begin{table*}[!tb]
    \centering
    \setlength{\tabcolsep}{1.2 mm}{
\resizebox{1.0\linewidth}{!}{
    \begin{tabular}{c|c|cc|cccl|ccccc}
    \hline
    CTRGCN && \multicolumn{2}{c|}{Threshold}  & \multicolumn{4}{c|}{Max} & \multicolumn{5}{c}{Mean} \\
    HDM05 &Accuracy& $l\ge3\uparrow$ & $\Delta B/B\ge5\%\uparrow$  & $l\uparrow$     & $\Delta a\uparrow$     & $\Delta\alpha\uparrow$    & $\Delta B/B\uparrow$    & $l\uparrow$   & $\Delta a\uparrow$  & $\Delta\alpha\uparrow$  & $\Delta B/B\uparrow$  & OM$\downarrow$  \\ \hline
    ST         &95.10\%& 5.0\% &1.7\%   &6.7  &0.28 & 1.78 &8.5\% & 0.67    &0.14 &0.31    &0.80\%    & 45.0\%   \\ \hline
    BEAT(Ours) &94.73\%& \textbf{9.9\%} &\textbf{3.2\%}   &6.0  &\textbf{0.37} & \textbf{3.39} &7.0\% & \textbf{0.79}    &0.14 &\textbf{0.38}    &\textbf{0.94\%}    & \textbf{44.4\%}   \\ \hline\hline
    NTU60 &Accuracy& $l\ge0.1\uparrow$ & $\Delta B/B\ge10\%\uparrow$  & $l\uparrow$     & $\Delta a\uparrow$     & $\Delta\alpha\uparrow$    & $\Delta B/B\uparrow$    & $l~\uparrow$   & $\Delta a\uparrow$  & $\Delta\alpha\uparrow$  & $\Delta B/B\uparrow$  & OM$\downarrow$  \\ \hline
    ST         &82.90\%& 14.0\% &19.4\%   &0.53  &0.12 & 0.31 &37.9\% & 0.05    &0.02 &0.03    &6.5\%    & 2.2\%   \\ \hline
    BEAT(Ours) &82.24\%& \textbf{18.9\%} &\textbf{22.6\%}   &\textbf{0.57}  &0.12 & \textbf{0.45} &\textbf{52.8\%} & \textbf{0.06}    &\textbf{0.03} &\textbf{0.04}    &\textbf{7.4\%}    & \textbf{0.9\%}   \\ \hline\hline
    NTU120 &Accuracy& $l\ge0.1\uparrow$ & $\Delta B/B\ge10\%\uparrow$  & $l~\uparrow$     & $\Delta a~\uparrow$     & $\Delta\alpha\uparrow$    & $\Delta B/B~\uparrow$    & $l~\uparrow$   & $\Delta a~\uparrow$  & $\Delta\alpha\uparrow$  & $\Delta B/B~\uparrow$  & OM$\downarrow$  \\ \hline
    ST         &74.59\%& 14.0\% &14.8\%   &0.53  &0.13 & 0.2 &34.0\% & 0.04    &0.019 &0.02    &5.4\%    & 1.7\%   \\ \hline
    BEAT(Ours) &74.25\%& \textbf{18.0\%} &\textbf{16.8\%}   &\textbf{1.54}  &\textbf{0.28} & 0.2 &\textbf{47.0\%} & \textbf{0.06}    &\textbf{0.022} &0.02    &\textbf{5.6\%}    & \textbf{1.0\%}   \\ \hline
    \end{tabular}}
}
    \caption{Defense results on CTRGCN. ST means standard training. Metrics are computed on the adversarial samples computed after ST training and BEAT training.}
    \label{tab:black_ctrgcn}
\end{table*}

\begin{table*}[!tb]
    \centering
    \setlength{\tabcolsep}{1.2 mm}{
\resizebox{1.0\linewidth}{!}{
    \begin{tabular}{c|c|cc|cccl|ccccc}
    \hline
    SGN && \multicolumn{2}{c|}{Threshold}  & \multicolumn{4}{c|}{Max} & \multicolumn{5}{c}{Mean} \\
    HDM05&Accuracy & $l\ge5\uparrow$ & $\Delta B/B\ge10\%\uparrow$  & $l\uparrow$     & $\Delta a\uparrow$     & $\Delta\alpha\uparrow$    & $\Delta B/B\uparrow$    & $l\uparrow$   & $\Delta a\uparrow$  & $\Delta\alpha\uparrow$  & $\Delta B/B\uparrow$  & OM$\downarrow$  \\ \hline
    ST         &94.16\%& 5.6\% &5.1\%   &20.5  &1.2 & 10.8 &22.3\% & 0.84    &0.05 &0.38    &1.1\%    & 88.8\%   \\ \hline
    BEAT(Ours) &93.78\%& \textbf{8.8\%} &\textbf{7.7\%}   &15.5  &\textbf{1.6} & 10.7 &\textbf{24.5\%} & \textbf{1.05}    &\textbf{0.07} &\textbf{0.56}    &\textbf{1.5\%}    & \textbf{87.9\%}   \\ \hline\hline
    NTU60 &Accuracy& $l\ge0.1\uparrow$ & $\Delta B/B\ge10\%\uparrow$  & $l\uparrow$     & $\Delta a\uparrow$     & $\Delta\alpha\uparrow$    & $\Delta B/B\uparrow$    & $l~\uparrow$   & $\Delta a\uparrow$  & $\Delta\alpha\uparrow$  & $\Delta B/B\uparrow$  & OM$\downarrow$  \\ \hline
    ST         &86.22\%& 13.7\% &6.8\%   &0.76 & 0.04 &0.13 &18.5\% & 0.06    &0.003 &0.01    &1.3\%    & 82.9\%   \\ \hline
    BEAT(Ours) &86.00\%& \textbf{16.8\%} &\textbf{8.9\%}   &\textbf{1.02} & \textbf{0.08} &\textbf{0.35} &\textbf{21.5\%} & \textbf{0.08}    &\textbf{0.004} &\textbf{0.02}    &\textbf{1.7\%}    & \textbf{75.6\%}   \\ \hline\hline
    NTU120 &Accuracy& $l\ge0.2\uparrow$ & $\Delta B/B\ge10\%\uparrow$  & $l~\uparrow$     & $\Delta a~\uparrow$     & $\Delta\alpha\uparrow$    & $\Delta B/B~\uparrow$    & $l~\uparrow$   & $\Delta a~\uparrow$  & $\Delta\alpha\uparrow$  & $\Delta B/B~\uparrow$  & OM$\downarrow$  \\ \hline
    ST         &74.15\%& 13.2\% &9.9\%   &1.25  &0.01 & 0.54 &40.2\%   &0.087 &0.005 &0.022   &2.3\%    & 75.4\%   \\ \hline
    BEAT(Ours) &73.54\%& \textbf{20.4\%} &\textbf{14.0\%}   &0.80  &\textbf{0.05} & 0.23 &37.7\%   &\textbf{0.103} &\textbf{0.006} &\textbf{0.023}   &\textbf{2.7\%}    & \textbf{73.1\%}   \\ \hline
    \end{tabular}}
}
    \caption{Defense results on SGN. ST means standard training. Metrics are computed on the adversarial samples computed after ST training and BEAT training.}
    \label{tab:black_sgn}
\end{table*}

\begin{table*}[!tb]
    \centering
    \setlength{\tabcolsep}{1.2 mm}{
\resizebox{1.0\linewidth}{!}{
    \begin{tabular}{c|c|cc|cccc|ccccc}
    \hline
    MSG3D && \multicolumn{2}{c|}{Threshold}  & \multicolumn{4}{c|}{Max} & \multicolumn{5}{c}{Mean} \\
    HDM05 &Accuracy& $l\ge1\uparrow$ & $\Delta B/B\ge10\%\uparrow$  & $l\uparrow$     & $\Delta a\uparrow$     & $\Delta\alpha\uparrow$    & $\Delta B/B\uparrow$    & $l\uparrow$   & $\Delta a\uparrow$  & $\Delta\alpha\uparrow$  & $\Delta B/B\uparrow$  & OM$\downarrow$  \\ \hline
    ST         &93.78\%& 0\% &0\%   &0.51  &0.25 & 0.91 &2.97\% & 0.20    &0.086 &0.23   &1.1\%    & 5.1\%   \\ \hline
    BEAT(Ours) &93.97\%& \textbf{3.3\%} &0\%   &\textbf{3.51}  &\textbf{0.32} & \textbf{11.53} &\textbf{3.56\%} & \textbf{0.28}    &\textbf{0.095} &\textbf{0.50}    &\textbf{1.2\%}    & \textbf{4.4\%}   \\ \hline\hline
    NTU60 &Accuracy& $l\ge0.1\uparrow$ & $\Delta B/B\ge15\%\uparrow$  & $l\uparrow$     & $\Delta a\uparrow$     & $\Delta\alpha\uparrow$    & $\Delta B/B\uparrow$    & $l~\uparrow$   & $\Delta a\uparrow$  & $\Delta\alpha\uparrow$  & $\Delta B/B\uparrow$  & OM$\downarrow$  \\ \hline
    ST         &89.36\%& 24.2\% &13.1\%   &0.57 & 0.17 &0.87 &53.8\% & 0.09    &0.03 &0.15    &8.9\%    & 2.0\%   \\ \hline
    BEAT(Ours) &89.16\%& \textbf{31.4\%} &\textbf{22.9\%}   &0.39 & \textbf{0.18} &0.79 &\textbf{75.2\%} & 0.09    &\textbf{0.04} &\textbf{0.18}    &\textbf{11.0\%}    & \textbf{0\%}   \\ \hline\hline
    NTU120 &Accuracy& $l\ge0.1\uparrow$ & $\Delta B/B\ge15\%\uparrow$  & $l~\uparrow$     & $\Delta a~\uparrow$     & $\Delta\alpha\uparrow$    & $\Delta B/B~\uparrow$    & $l~\uparrow$   & $\Delta a~\uparrow$  & $\Delta\alpha\uparrow$  & $\Delta B/B~\uparrow$  & OM$\downarrow$  \\ \hline
    ST         &84.71\%& 14.3\% &6.3\%   &0.29  &0.10 & 0.50 &34.3\%   &0.06 &0.02 &0.11   &6.8\%    & 0\%   \\ \hline
    BEAT(Ours) &82.70\%& \textbf{24.6\%} &\textbf{23.2\%}   &\textbf{0.51}  &\textbf{0.17} & \textbf{0.71} &\textbf{60.0\%}   &\textbf{0.08} &\textbf{0.03} &\textbf{0.15}   &\textbf{9.0\%}    & 0\%   \\ \hline
    \end{tabular}}
}
    \caption{Defense results on MSG3D. ST means standard training. Metrics are computed on the adversarial samples computed after ST training and BEAT training.}
    \label{tab:black_msg3d}
\end{table*}

\section{BEAT Inference}
For sake of self-containedness, we first give all key equations in our model. We model the joint clean-adversarial distribution:
\begin{equation}
\label{eq:fullJointEnergy}
    p_{\theta}(\mathbf{x}, \Tilde{\mathbf{x}}, y) = \frac{exp\{g_{\theta}(\mathbf{x})[y] + g_{\theta}(\mathbf{\Tilde{\mathbf{x}}})[y] - \lambda d(\mathbf{x}, \Tilde{\mathbf{x}})\}}{Z(\theta)}
\end{equation}
where $\mathbf{x}$ and $\Tilde{\mathbf{x}}$ are data samples and their corresponding adversarial samples under class $y$. $\lambda$ is a weight and $d(\mathbf{x}, \Tilde{\mathbf{x}})$ measures the distance between clean samples and their corresponding adversarial samples. $g_{\theta}$ is a given black-box classifier. Maximizing the log-likelihood of the joint probability gives:
\begin{align}
\label{eq:logLikelihood}
    log\ &p_{\theta}(\mathbf{x}, \Tilde{\mathbf{x}}, y) = log\  p_{\theta}(\Tilde{\mathbf{x}} | \mathbf{x}, y) + log\ p_{\theta}(\mathbf{x}, y) \nonumber \\
    &= log\  p_{\theta}(\Tilde{\mathbf{x}} | \mathbf{x}, y) + log\ p_{\theta}(y|\mathbf{x}) + log\ p_{\theta}(\mathbf{x}) 
\end{align}
$log\ p_{\theta}(y|\mathbf{x})$ is simply a classification likelihood. Both $p_{\theta}(\mathbf{x})$ and $p_{\theta}(\Tilde{\mathbf{x}} | \mathbf{x}, y)$ are intractable, so sampling is needed. We keep the pre-trained classifier intact and append a model with parameters $\theta'$ behind the recognizer using a skip connection: logits = $f_{\theta'}(\phi(x)) + g_\theta(x)$. Here, $\phi(x)$ is the original logits $\phi(x) = g_\theta(x)$ to keep the \textit{black-boxness} of BEAT. $f_{\theta'}$ is a two-layer MLP network, each layer with a dimension being the same as the class number in the data. The BEAT model then can be represented as:
\begin{align}
\label{eq:ptBayesianClassifier}
    p(y' | x', \mathbf{x}, \Tilde{\mathbf{x}}, y) = E_{\theta'\sim p(\theta')}[p(y' | x', \mathbf{x}, \Tilde{\mathbf{x}}, y, \theta, \theta')] \nonumber \\
    = \int p(y' | x', \theta') p(\theta' | \mathbf{x}, \Tilde{\mathbf{x}}, y, \theta) d\theta' \nonumber \\
    \approx \frac{1}{N}\sum_{i = 1}^Np(y'|x',\theta'_i), \theta' \sim p(\theta' | \mathbf{x}, \Tilde{\mathbf{x}}, y, \theta)
\end{align}
We assume $\theta$ is obtained through pre-training, then:
\begin{align}
\label{eq:ptModelSampling}
    &\{\mathbf{x}, \Tilde{\mathbf{x}}, y\}_t | \theta, \theta'_{t-1} \sim p(\mathbf{x}, \Tilde{\mathbf{x}}, y | \theta, \theta'_{t-1}) \nonumber \\
    &\theta'_t | \{\mathbf{x}, \Tilde{\mathbf{x}}, y, \theta\}_t \sim p(\theta' | \{\mathbf{x}, \Tilde{\mathbf{x}}, y\}_t, \theta)
\end{align}
where the inference on $\theta'_t$ can be computed via sampling.
\subsection{Inference details}
\cref{eq:logLikelihood} maximizes the log likelihood of the joint probability, which needs to compute three gradients $\frac{\partial log p_{\theta}(\Tilde{\mathbf{x}} | \mathbf{x}, y)}{\partial\theta}$, $\frac{\partial log p_{\theta}(y|\mathbf{x})}{\partial\theta}$ and $\frac{\partial log p_{\theta}(\mathbf{x})}{\partial\theta}$. Instead of directly maximizing $log p_{\theta}(y|\mathbf{x})$, we minimize the \textit{cross-entropy} on the logits, so $\frac{\partial log p_{\theta}(y|\mathbf{x})}{\partial\theta}$ is straightforward. Next, $\frac{\partial log p_{\theta}(\mathbf{x})}{\partial\theta}$ can be approximated by~\cite{nijkamp_anatomy_2019}:
\begin{align}
\label{eq:pxgradient}
    \frac{\partial log p_{\theta}(\mathbf{x})}{\partial\theta} \approx \frac{\partial}{\partial\theta}[\frac{1}{L_1}\sum_{i=1}^{L_1}U(g_{\theta}(\mathbf{x}^+_i)) - \frac{1}{L_2}\sum_{i=1}^{L_2}U(g_{\theta}(\mathbf{x}^-_i))]
\end{align}
where $U$ gives the mean over the logits, \{$\mathbf{x}^+_i$\}$_{i=1}^{L_1}$ are a batch of training samples and \{$\mathbf{x}^-_i$\}$_{i=1}^{L_2}$ are i.i.d. samples from $p_{\theta}(\mathbf{x})$ via Stochastic Gradient Langevin Dynamics (SGLD)~\cite{welling_bayesian_2011}:
\begin{align}
\label{eq:SGLD_x}
    \mathbf{x}^-_{t+1} = \mathbf{x}^-_{t} + \frac{\epsilon^2}{2} \frac{\partial log p_{\theta}(\mathbf{x}^-_t)}{\partial \mathbf{x}^-_t} + \epsilon E_t, \epsilon > 0, E_t\in \mathbf{N}(0, \mathbf{I}) 
\end{align}
where $\epsilon$ is a step size, $\mathbf{N}$ is a Normal distribution and $\mathbf{I}$ is an identity matrix. Similarly for $\frac{\partial log p_{\theta}(\Tilde{\mathbf{x}} | \mathbf{x}, y)}{\partial\theta}$:
\begin{align}
\label{eq:pxtildegradient}
    \frac{\partial log p_{\theta}(\Tilde{\mathbf{x}}|\mathbf{x}, y)}{\partial\theta} = \frac{\partial}{\partial\theta}\{g_{\theta}(\mathbf{\Tilde{\mathbf{x}}})[y] - \lambda d(\mathbf{x}, \Tilde{\mathbf{x}})\} 
\end{align}
where $\Tilde{\mathbf{x}}$ can be sampled via:
\begin{align}
\label{eq:SGLD_xtilde}
    \Tilde{\mathbf{x}}_{t+1} = \Tilde{\mathbf{x}}_{t} + \frac{\epsilon^2}{2} \frac{\partial log p_{\theta}(\Tilde{\mathbf{x}}_t| \mathbf{x}, y)}{\partial\Tilde{\mathbf{x}}_t} + \epsilon E_t, \epsilon > 0, E_t\in \mathbf{N}(0, \mathbf{I})
\end{align}
Further, instead of naive SGLD, we use Persistent Contrastive Divergence~\cite{tieleman_training_2008} with a random start~\cite{du_implicit_2020}. With \cref{eq:pxgradient}-\cref{eq:SGLD_xtilde}, we use a stochastic gradient optimizer to solve \cref{eq:logLikelihood}. Besides \cref{eq:logLikelihood}, BEAT also needs to sample $\theta'$.  Stochastic Gradient Hamiltonian Monte Carlo~\cite{chen_stochastic_2014} is a common strategy for such sampling tasks. However, we find that it cannot efficiently explore the target density due to the high correlations between parameters in $\theta'$. Therefore, we use Stochastic Gradient Adaptive Hamiltonian Monte Carlo~\cite{springenberg_bayesian_2016}:

\begin{align}
\label{eq:SGAHMC}
&\theta'_{t+1} = \theta'_t - \sigma^2\mathbf{C}^{-1/2}_{\theta'_t}\mathbf{h}_{\theta'_t} + \mathbf{N}(0, 2F\sigma^3\mathbf{C}^{-1}_{\theta'_t} - \sigma^4\mathbf{I}) \nonumber\\
&\mathbf{C}_{\theta'_t} \leftarrow (1 - \tau^{-1})\mathbf{C}_{\theta'_t} + \tau^{-1}\mathbf{h}_{\theta'_t}^2
\end{align}
where $\sigma$ is the step size, $F$ is called friction coefficient, $\mathbf{h}$ is the stochastic gradient of the system, $\mathbf{N}$ is a Normal distribution and $\mathbf{I}$ is an identity matrix, $\mathbf{C}$ is a pre-conditioner and updated by an exponential moving average and $\tau$ is chosen automatically~\cite{springenberg_bayesian_2016}. The BEAT inference is detailed in \cref{alg:BEAT}.

\begin{algorithm}[tb]
\SetAlgoLined
\textbf{Input}: $\mathbf{x}$: training data; $N_{tra}$: the number of training iterations; $M_1$ and $M_2$: sampling iterations; $M_{\theta'}$: sampling iterations for $\theta'$; $f_{\theta'}$: appended models with parameter $\{\theta'_1, \dots, \theta'_N\}$; $N$: the number of appended models\;
\textbf{Output}: $\{\theta'_1, \dots, \theta'_N\}$: appended network weights\;
\textbf{Init}: randomly initialize $\{\theta'_1, \dots, \theta'_N\}$\;

\For{i = 1 to $N_{tra}$}{
    \For{n = 1 to $N$}{
        Randomly sample a mini-batch data $\{\mathbf{x}, y\}_i$\;
        Compute $h_1 = \frac{\partial log p_{\theta'}(y|\mathbf{x})}{\partial\theta'}$\;
        Obtain $\mathbf{x}_0$ via random noise~\cite{du_implicit_2020}\;
        \For{t = 1 to $M_1$}{
            Sample $\mathbf{x}_t$ from $\mathbf{x}_{t-1}$ via \cref{eq:SGLD_x}\;
        }
        Compute $h_2 = \frac{\partial log p_{\theta'}(\mathbf{x})}{\partial\theta'}$ via \cref{eq:pxgradient}\;
        Obtain $\Tilde{\mathbf{x}}_0$ from $\mathbf{x}_i$ with a perturbation\;
        \For{t = 1 to $M_2$}{
            Sample $\Tilde{\mathbf{x}}_t$ from $\Tilde{\mathbf{x}}_{t-1}$ via \cref{eq:SGLD_xtilde}\;
        }
        Compute $h_3 = \frac{\partial log p_{\theta'}(\Tilde{\mathbf{x}} | \mathbf{x}, y)}{\partial\theta'}$ via \cref{eq:pxtildegradient}\;
        $\mathbf{h}_{\theta'} = h_1 + h_2 + h_3$\;
    
        \For{t = 1 to $M_{\theta'}$}{
        Update $\theta'_n$ with $\mathbf{h}_{\theta'}$ via \cref{eq:SGAHMC}\;
        }
    }
}
\Return $\{\theta'_1, \dots, \theta'_N\}$\;
\caption{Inference on BEAT}
\label{alg:BEAT}
\end{algorithm}

\section{Connections to Existing Defense Methods}

BEAT has intrinsic connections with Adversarial Training (AT) and Randomised smoothing (RS). Since the potential attacker is unknown \textit{a priori}, $d(\mathbf{x}, \Tilde{\mathbf{x}})$ in Eq.~\ref{eq:fullJointEnergy} needs to capture the full adversarial distribution. In this sense, $d(\mathbf{x}, \Tilde{\mathbf{x}})$ can be seen as a generalization of AT and RS. AT optimizes~\cite{madry_towards_2018}:
\begin{equation}
\min_{\theta} E_{\mathbf x}[\max_{\sigma\in S} L(\theta, \mathbf{x} + \sigma, y)]
\end{equation}
where $L$ is the classification loss function, $\sigma$ is a perturbation from a set $S$ which is normally constrained within a ball and $\sigma$ needs to be computed by a pre-defined attacker. Here, $d(\mathbf{x}, \Tilde{\mathbf{x}})$ is the Euclidean distance between $\mathbf{x}$ and $\Tilde{\mathbf{x}}$ within the ball $S$. In RS, the robust classifier is obtained through~\cite{Cohen19_ICML}:
\begin{equation}
    \argmax_{y \in \mathbf{y}} p(g_\theta(\mathbf{x} + \sigma)=y) \text{ where } \sigma \sim \mathbf{N}
\end{equation}
where a perturbation $\sigma$ is drawn from an isotropic Gaussian. $d(\mathbf{x}, \Tilde{\mathbf{x}})$ essentially plays the role of the Gaussian to describe the perturbation distribution. However, neither AT nor RS can capture the fine-grained structure of the adversarial distribution, because AT merely uses the most aggressive adversarial example, and RS often employs simple \textit{isotropic} distributions (e.g. Gaussian/Laplacian)~\cite{Zhang_black_2020}. Therefore, we argue $d(\mathbf{x}, \Tilde{\mathbf{x}})$ should be data/task specific and should not be restricted to isotropic forms. This is because adversarial samples are near the data manifold, both on-manifold and off-manifold~\cite{diao_basarblack-box_2021}, so the data manifold geometry should dictate the parameterization of $d(\mathbf{x}, \Tilde{\mathbf{x}})$. In general, there are two possible avenues to model the data manifold: implicit and explicit. Explicit formulations can be used if it is relatively straightforward to parameterize the manifold geometry; or a data-driven model can be used to implicitly learn the manifold. Either way, the manifold can then be devised with a distance function to instantiate $d(\mathbf{x}, \Tilde{\mathbf{x}})$. Our current paper focuses on skeletal HAR problem, so $d(\mathbf{x}, \Tilde{\mathbf{x}})$ is explicitly parameterized by motion dynamics and bone length. Other forms are also possible for other data types but we will leave it to future work. 

\section{Experimental Settings}
In this section, we give more details regarding our experimental settings.
\paragraph{Datasets:} We choose three widely adopted benchmark datasets in HAR: HDM05~ \cite{muller_documentation_2007}, NTU60~\cite{shahroudy_ntu_2016} and NTU120~\cite{liu_ntu_2020}. HDM05 dataset has 130 action classes, 2337 sequences from 5 non-professional actors. NTU60 and NTU120 have two settings: cross-subject and cross-view. We only use the cross-subject data. NTU60 includes 56568 skeleton sequences with 60 action classes, performed by 40 subjects. NTU120 extends NTU60 with an additional 57,367 skeleton sequences over 60 extra action classes, totalling 113,945 samples over 120 classes captured from 106 distinct subjects. We follow the protocols in \cite{wang_understanding_2021,diao_basarblack-box_2021} for pre-processing. 

\paragraph{Detailed Attack Setting:} Here we report the detailed robustness evaluation. To test the classifier robustness, we employ state-of-the-art attackers designed for skeleton-based HAR: SMART ~\cite{wang_understanding_2021}, CIASA~\cite{liu_adversarial_2020} and BASAR~\cite{diao_basarblack-box_2021}, and follow their default settings. Further, we use the untargeted attack, which is the most aggressive setting. SMART is the $l_2$ norm-based white-box attack, whose learning rate is set as 0.005. CIASA is the $l_{\infty}$ norm-based white-box attack, whose learning rate is set as 0.005, and the perturbation budgets define different clipping strengths for different joint. Following their paper, the variables are set to 0.01, 0.05, 0.15, 0.25 for the joint hips, knees, ankles, and feet, respectively. Since all attackers are iterative approaches and more iterations lead to more aggressive attacks, we evaluate the defenders with 1000-iteration SMART attack on all datasets, CIASA-1000 on HDM05 and CIASA-100 on NTU 60/120 since running CIASA-1000 on NTU 60/120 is prohibitively slow (approximately 2 months on a Nvidia Titan GPU). BASAR is the $l_2$ norm-based black-box threat model. We use the same iterations as in their paper, i.e. 500 iterations on HDM05 and 1000 iterations on NTU 60/NTU 120. After training, we collect the correctly classified testing samples for attack. 

\paragraph{Training details of BEAT:}
\label{setting}
In SGLD in \cref{eq:SGLD_x} and \cref{eq:SGLD_xtilde}, instead of using an identity matrix as the covariance matrix for $E_t$, we use $\sigma^2\mathbf{I}$ where $\sigma = 0.005$. In \cref{alg:BEAT}, we set $M_1 = 10$, $M_2 = 10$ and $b=0.05$. In \cref{eq:pxtildegradient}, we use $\lambda=10^{-3}$. In \cref{eq:SGAHMC}, $\tau$ is automatically chosen~\cite{springenberg_bayesian_2016}. We set $F=10^{-5}$. We also run \cref{eq:SGAHMC} $M_{\theta'}=30$ steps each time for sampling.  One thing to notice is that in Line 13-15 in \cref{alg:BEAT}, only one $\Tilde{x}$ is sampled in each iteration. To model $p_{\theta}(\Tilde{\mathbf{x}} | \mathbf{x}, y)$, multiple $\Tilde{x}$ can indeed be sampled. In practice, we find one sample in every iteration can give good performances, because in different iterations different $\Tilde{x}$s are sampled. 

In practice, we weight the $h_1,h_2,h_3$ in \cref{alg:BEAT}:
\begin{align}
\label{eq:weights_h}
    \mathbf{h}_{\theta'}/\mathbf{h} = w_1 h_1 + w_2 h_2 + w_3 h_3
\end{align}

The weights in \cref{eq:weights_h}, $\epsilon$ in \cref{eq:SGLD_x} and \cref{eq:SGLD_xtilde}, and $\sigma$ in \cref{eq:SGAHMC} can be grouped into several universal settings according to different datasets. Unless specified otherwise, we use $w_1=1$, $w_2=0.3$, $w_3=0.1$, $\epsilon=0.01$ by default. On HDM05, $\sigma=0.001$ for STGCN and MSG3D, $\sigma=0.01$ for CTRGCN and $\sigma=0.005$ for SGN. On NTU60, $w_1=1$,$w_2=0.1$,$w_3=0.05$ and $\sigma=0.002$ for CTRGCN, and $\sigma=0.0015$ for MSG3D. On NTU60 and NTU120, $\sigma=0.005$ for SGN and $w_1=1$,$w_2=0.01$,$w_3=0.005$, $\epsilon=2$ and $\sigma=0.5$ for STGCN. On NTU120, $\sigma=0.0015$ for CTRGCN and $\sigma=10^{-4}$ for MSG3D.

Given that decision-based attack tends to use bigger perturbations, we increase $w_3$ in \cref{{eq:weights_h}} and $b$ in \cref{alg:BEAT} when defending against decision-based attack. Unless specified otherwise, we use $w_1=1$, $w_2=0.1$, $w_3=1$, $b=0.5$, $\sigma=0.02$,$M_1 = 2$ and $M_2 = 2$ by default in defense against black-box attack. The decreasing of sampling iterations is to increase the randomness of $\Tilde{\mathbf{x}}$. For STGCN on NTU60, we use $b=1$. For CTRGCN on NTU60 and NTU120, we use $w_1=1$, $w_2=0.01$, $w_3=1$ and $b=1$. For SGN on NTU120, we set $w_1=1$, $w_2=0.01$, $w_3=10$ and $b=1$.

\paragraph{Random Smoothing setting:}  
For RS, we follow~\cite{zheng_towards_2020} and the defense strategy is:
\begin{equation}
    \argmax_{y \in \mathbf{y}} p(g_\theta(\mathbf{\widehat{x}})=y)
\label{eq:gaussian}
\end{equation}
where $y\in\mathbf{y}$ is the class label. $\mathbf{\widehat{x}}$ is the noisy samples, generated by adding a Gaussian noise to the original sample then temporal-filtering it by a $1 \times 5$ Gaussian kernel. The Gaussian noise is drawn from $\mathcal{N}(0, \delta^2 \mathbf{I})$ where $\delta$ is the standard deviation, $\mathbf{I}$ is an identity matrix. We set $\delta=0.5$ on HDM05. Our preliminary experiments show setting $\delta=0.5$ on NTU 60/120 severely compromise the accuracy, so we set $\delta=0.1$ on both NTU60 and NTU120.


%
%

\bibliography{appendix}